\documentclass{ieeeaccess}
\usepackage{amsmath,amssymb,amsfonts}
\usepackage{graphicx}
\usepackage{textcomp}
\def\BibTeX{{\rm B\kern-.05em{\sc i\kern-.025em b}\kern-.08em
    T\kern-.1667em\lower.7ex\hbox{E}\kern-.125emX}}

\usepackage{algorithm}
\usepackage{algpseudocode}
\usepackage{amsmath}
\DeclareMathOperator*{\argmax}{argmax}
\usepackage{booktabs}

\usepackage{biblatex}

\usepackage{placeins}
\newtheorem{lemma}{Lemma}
\usepackage{caption}
\begin{document}
\history{Date of publication xxxx 00, 0000, date of current version xxxx 00, 0000.}
\doi{10.1109/ACCESS.2026.3685576}

\title{Best Agent Identification for General Game Playing}
\author{\uppercase{Matthew Stephenson}\authorrefmark{1}, \IEEEmembership{Member, IEEE}, \uppercase{Alex Newcombe}\authorrefmark{1}, \uppercase{{\'E}ric Piette}\authorrefmark{2}, \uppercase{and Dennis J.N.J. Soemers}\authorrefmark{3}, \IEEEmembership{Member, IEEE}}
\address[1]{College of Science and Engineering, Flinders University, Adelaide, South Australia, Australia}
\address[2]{Information and Communication Technologies, Electronics, and Applied Mathematics Institute, UCLouvain, Louvain-la-Neuve, Belgium}
\address[3]{Department of Advanced Computing Sciences, Maastricht University, Maastricht, the Netherlands}

\markboth
{Stephenson \headeretal: Best Agent Identification for General Game Playing}
{Stephenson \headeretal: Best Agent Identification for General Game Playing}

\corresp{Corresponding author: Matthew Stephenson (e-mail: matthew.stephenson@flinders.edu.au).}

\begin{abstract}
We present an efficient and generalised procedure to accurately identify the best (or near best) performing algorithm for each sub-task in a multi-problem domain. Our approach treats this as a set of best arm identification problems for multi-armed bandits, where each bandit corresponds to a specific task and each arm corresponds to a specific algorithm or agent. We propose an optimistic selection process based on a chosen confidence interval, that ranks each arm across all bandits in terms of their potential to influence our overall simple regret. We evaluate the performance of our approach on two of the most popular general game playing domains, the General Video Game AI (GVGAI) framework and the Ludii general game playing system, with the goal of selecting a high-performing agent for each game using a limited number of available trials. Compared to previous best arm identification algorithms for multi-armed bandits, our results demonstrate a substantial performance improvement in terms of average simple regret and average probability of error. This novel approach can be used to significantly improve the quality and accuracy of agent evaluation procedures for general game frameworks, as well as other multi-task domains with high algorithm runtimes.
\end{abstract}


\titlepgskip=-15pt

\maketitle
\section{Introduction}

The field of general game playing focuses on the development of AI agents that are able to effectively play a wide variety of games \cite{Genesereth_Love_Pell_2005}. This can often be viewed as a multi-problem domain, where each agent needs to perform well across a collection of individual tasks. Designing artificial general agents that are able to successfully play a wide range of games has long been a core research area for artificial intelligence and machine learning, leading to significant breakthroughs in the areas of deep reinforcement learning \cite{Torrado2018DeepRL,Goldwaser_Thielscher_2020}, search and planning \cite{7860383,7317937}, transfer learning \cite{10.5555/1625275.1625383, 2eee97e2f1ce42508011c549376c2461}, and evolutionary algorithms \cite{8080420}. 
However, as the number of games and agents developed for these frameworks increases, so too does the evaluation time needed to reliably assess the performance of each agent. Many agents do not play identically when repeatedly attempting a game, often requiring a large number of trials to obtain an accurate performance estimate \cite{Bontrager_Khalifa_Mendes_Togelius_2021}. While there is often ``no free lunch'' when it comes to producing more results in less time \cite{10.1109/CIG.2017.8080410}, several approaches have been proposed to reduce the number of unnecessary evaluations using prior domain knowledge \cite{10.1109/CEC48606.2020.9185834, 10.5555/3618408.3618428}.

Rather than needing to accurately assess the game playing abilities of each agent across all games, our aim is to develop a domain agnostic approach that can efficiently identify the best (or near best) performing algorithm (i.e., agent) for each task (i.e., game) within a defined set. Identifying a high-performing agent for each game can be useful for several purposes, such as training prediction models for portfolio/ensemble agents or detecting skill gaps in the existing agent suite \cite{8848089, AIIDE1715828}. We frame this challenge as a multi-bandit best arm identification problem \cite{NIPS2011_c4851e8e}, where each bandit represents a specific game and each arm represents an available agent for playing that game. In this comparative setup, our goal is to select a single agent for each of our available games, with the goal of maximising the total reward (i.e., number of wins) across all games. To help us make our selection, we can run a limited number of exploratory trials to estimate the performance of each game-agent pair. The challenge therefore lies in determining which game-agent pairs (i.e., trials) should be run to help identify the top performing agent for each game.

In this paper, we propose a novel multi-bandit best arm identification algorithm called RCP (Regret Change Potential). This approach combines an ``optimism in the face of uncertainty'' philosophy with a chosen confidence interval estimate, selecting arms that have the highest potential for minimising the overall simple regret across all bandits.
We evaluate and compare the performance of our RCP approach with multiple confidence interval estimates, against several prior multi-bandit best arm identification algorithms for two of the most popular general game playing domains. This includes the General Video Game AI (GVGAI) framework \cite{gvgaibook2019} and the Ludii general game playing system \cite{23abe395562246cf89bbe73e83d19717}, both of which are frequently used within the academic community for developing and comparing general game playing techniques \cite{Perez-Liebana_Samothrakis_Togelius_Schaul_Lucas_2016,84079a9259574ebb8f5b60771e5d02e7}. Our results demonstrate that our proposed RCP approach provides significant performance improvements with regards to best agent identification, when compared to alternative state-of-the art techniques.

The remainder of this paper is organised as follows: Section 2 formalises the multi-bandit best arm identification problem and its applicability to best agent identification for general game playing. Section 3 describes relevant prior work on general game systems, efficient agent evaluation procedures, best arm identification algorithms, and various confidence interval estimates. Section 4 defines our proposed RCP approach, and the motivation behind its design. Section 5 provides details on our experimental setup using the GVGAI and Ludii general game playing domains. Section 6 presents and discusses the relative performance of each best arm identification algorithm. Section 7 summarises our results, highlighting potential areas for further investigation and improvement.

\section{Problem Setup}
\label{sec:Problem Setup}

In this section, we provide a formalisation of the multi-bandit best arm identification problem and its applicability to our intended domain of general game playing. 

\subsection{Multi-Bandit Best Arm Identification}

Best-arm identification is a variation of the more traditional stochastic multi-armed bandit problem. Rather than maximising the cumulative reward over multiple arm pulls, the aim of best-arm identification is to maximise the reward of a single arm pull after a defined exploration period (i.e., select the arm with the highest expected reward after a limited number of prior arm pulls). 
This problem can be expanded further into the multi-bandit best arm identification problem, where instead of a single bandit there are now multiple bandits available to choose from, each with their own selection of arms and reward distributions. Rather than identifying the single best arm, the goal of multi-bandit best arm identification is to identify the best performing arm for each individual bandit. Once the defined exploration period ends, a single arm pull of each bandit is performed and the rewards from each pull are summed to give the final result. The challenge of the multi-bandit setup comes from the need to effectively distribute the available exploratory pulls across all bandits, as well as the arms each bandit has, to obtain the necessary information to maximise the combined reward from a single pull of each bandit.
A more formalised definition of the multi-bandit best arm identification problem is provided below, with the majority of terms borrowed directly from \cite{NIPS2011_c4851e8e}.

The setup for a multi-bandit environment can be defined as follows.
Let $M$ be the number of bandits and $K$ represents the number of arms available for each bandit. Each individual bandit-arm pair $(m,k)$ can be characterised by a stationary reward distribution $\nu_{mk}$, bounded in $[0,b]$ and with mean reward $\mu_{mk}$. Assuming that each bandit has a unique best arm, $k_{m}^\star$ and $\mu_{m}^\star$ represent the index and mean of the best arm for bandit $m$ (i.e., $k_m^\star = \argmax_{1\leq k\leq K} \mu_{mk}, \ \mu_m^\star = \max_{1\leq k\leq K} \mu_{mk}$).

The problem of Best Arm Identification occurs within a multi-bandit environment when the true reward distributions $\{\nu_{mk}\}$ of our bandits are unknown, and is often framed as a prediction task. During each round $t = 1 \dots n$, we can pull a single bandit-arm pair $I(t) = (m,k)$ and observe an independent sample reward drawn from the distribution $\nu_{I(t)}$. Let $T_{mk}(t)$ be the number of times that bandit-arm pair $(m,k)$ has been pulled by the end of round $t$, then the mean reward of this bandit-arm pair can be estimated as 
$\hat{\mu}_{mk}(t) = \frac{1}{T_{mk}(t)} \sum_{s=1}^{T_{mk}(t)} X_{mk}(s)$, where $X_{mk}(s)$ is the $s$-th sample observed from $\nu_{mk}$. 
After all $n$ rounds have finished, for each bandit $m$ we return the arm $J_{m}(n) = \argmax_{k} \hat{\mu}_{mk}(n)$ with the highest estimated mean reward. 

To evaluate the performance of our approach after $n$ rounds, for each bandit we compare the expected reward for our selected arm against that of the true best arm. This comparison is usually done by calculating the average difference in expected rewards for each arm pair, known as the simple regret $r(n) = \frac{1}{M}\sum_{m=1}^M(\mu_m^\star - \mu_{mJ_m(n)})$.
In certain situations, the percentage of incorrectly identified best arms may be a more desirable measure of performance, $e(n) = \frac{1}{M}\sum_{m=1}^M\mathbb{P}(J_m(n) \neq k_{m}^\star)$. For our case study, we primarily focus on using the average simple regret $r(n)$ as our measure of algorithm performance, although additional performance comparisons using the average probability of error measure $e(n)$ are provided as well.

\subsection{Best Agent Identification for General Game Playing}
\label{Best Agent Identification for General Game Playing}

The above problem of multi-bandit best arm identification is analogous to determining the best performing algorithms across a variety of different tasks. Each task can be treated as a single multi-armed bandit, with each arm representing a potential algorithm that can perform this task (albeit with varying degrees of success). 
Using this framing allows us to represent our task of identifying the best agent for each game within a general game framework as a multi-bandit best arm identification problem. Each game is treated as its own multi-arm bandit, with each arm of this bandit representing a different game playing agent. A pull of a specific arm for a specific bandit corresponds to running a single trial for the associated game-agent pair, with the agent's outcome at the end of this trial determining the reward obtained. 
While prior approaches for solving multi-bandit best arm identification problems have been proposed for the general case described above, our specific focus on general game playing introduces several additional domain considerations.

\subsubsection{Reward Bounds}
One assumption we can make for the domain of general game playing is that the outcome for an agent at the end of a trial is always between 0 and 1 (inclusive). Typically an outcome of 0 indicates a loss and an outcome of 1 indicates a win, with other outcomes between these indicating either a draw or positional ranking among multiple players. This means that the value of $b$ in our reward distribution bounds is always equal to 1 (i.e., all rewards bounded in $[0,1]$). 

\subsubsection{Unequal Number of Arms}
Another consideration is that the above definition of a multi-bandit environment  assumes an equal number of arms $K$ for each bandit. However, in our application to general game playing it is often the case that certain agents may not be able to play every game available. In these situations, we can simply apply a generalisation to the above definitions that replaces the static value $K$ with a bandit-specific value $K_m$ that represents the number of arms available for a given bandit $m$.

\subsubsection{Multiple Best Arms}
Lastly, when comparing multiple agents for a single game, the assumption that there is always a unique best arm $k_{m}^\star$ for any bandit $m$ is unlikely to hold. For example, any game where several agents are always able to win results in an equally high reward estimation of 1.0 for multiple arms. To account for this, we define our $\argmax$ function to always return a single best argument combination, chosen at random from the highest value arguments. We also need to modify the average probability of error measure $e(n)$, so that selecting any arm with the highest expected reward $\mu_{m}^\star$ is recorded as a correct prediction (i.e., $e(n) = \frac{1}{M}\sum_{m=1}^M\mathbb{P}(\mu_m^\star \neq \mu_{mJ_m(n)})$).

\section{Background}
\label{sec:Background}

\subsection{General Game Systems}

While many different general game playing domains have been proposed over the past decade, arguably the two most popular for academic research have been the General Video Game AI framework (GVGAI) and the Ludii general game playing system.

\subsubsection {General Video Game AI Framework}

The General Video Game AI (GVGAI) framework contains over 100 simple arcade-style video games, described using the Video Game Description Language (VGDL) \cite{gvgaibook2019}. The GVGAI framework has been the subject of a long running general game playing competition \cite{Perez-Liebana_Samothrakis_Togelius_Schaul_Lucas_2016}, leading to several dozen agents being developed for it over the years \cite{7860430, 7860448, 8080420, Weinstein_Littman_2012, 7860398}. While there are several auxiliary research tracks available for the GVGAI framework, in this paper we focus solely on the games and agents developed for the main single-player planning track. Each game includes five distinct levels, which all use the same base set of mechanics but vary the number and initial locations of objects. Previous analysis on agent performance across the corpus of GVGAI games has observed that agents often produce highly varied reward distributions \cite{Bontrager_Khalifa_Mendes_Togelius_2021}. Many of the games available within the GVGAI framework also have some form of stochasticity, such as randomised items or enemy behaviour, making agent evaluation an inherently noisy process. These factors, along with the time consuming nature of running dozens of trials across thousands of game-agent pairs, makes the GVGAI framework an ideal case study for best agent identification.

\subsubsection {Ludii general game playing system}
In contrast to the GVGAI framework's focus on video games, the Ludii general game playing system instead provides over 1000 board and puzzle games \cite{23abe395562246cf89bbe73e83d19717}. This makes Ludii one of the largest collections of playable games for AI research purposes. Games within Ludii are described using the ludemic game description language, which supports a wide range of mechanics and game types. This includes stochastic or hidden information, alternating or simultaneous moves, and player counts ranging from 1 to 16 \cite{8847949}. Ludii has also been used as a competition framework for promoting the development of general game playing agents \cite{84079a9259574ebb8f5b60771e5d02e7}, as well as research into general game heuristics \cite{9619052}. Much like the GVGAI framework, agent evaluation in Ludii is a lengthy and imperfect process, often requiring maximum turn limits to even guarantee that a game ends. These aspects make Ludii a prime candidate for applying more efficient best agent identification algorithms.

\subsection{Efficient Agent Evaluation}

While we believe that this is the first time the task of best agent identification for general game systems has been framed from a multi-arm bandit perspective, this section discusses alternative techniques for more efficiently evaluating general game agents.

\subsubsection{Game Subsets}
Rather than evaluating all agents across the full suite of available games, one alternative is to identify a representative subset of games to evaluate on. This approach has been previously investigated for both the General Video Game AI framework \cite{10.1109/CEC48606.2020.9185834} and Arcade Learning Environment \cite{10.5555/3618408.3618428}. This technique has also been mentioned as possible future work for the Ludii general game playing system \cite{9618990,10.1007/978-3-031-34017-8_11} but has yet to be properly investigated. While this approach has a similar motivation to our current problem, being that testing all game-agent pairs a sufficient number of times to get a reliable indication of performance is computationally expensive, our goal is to instead focus on identifying the best agent for each individual game, rather than assessing the general performance of each agent. These approaches also use domain-specific measures of game and/or agent similarity to identify a representative game subset, whereas our presented approach requires no such domain knowledge.

\subsubsection{Active Learning}
Active Learning is a branch of Machine Learning where labelled training instances are selected through an intelligent process \cite{pmlr-v16-settles11a}. This typically occurs when there is a large quantity of unlabelled instances, but labelling any given instance is costly to perform. An active learning algorithm attempts to identify which instances should be labelled to provide maximal training benefit.
This situation is highly similar to our described problem, but instead of an abundance of unlabelled instances that need labelling we instead have a noisy and/or probabilistic labelling process. In both cases, the core motivation that obtaining a large number of accurate instance labels is prohibitively expensive remains the same. While previous research has demonstrated that active learning approaches can be used to minimise reward estimate uncertainty for multi-armed bandits \cite{5967348}, the problem of best arm identification has not yet been explored.

\subsection{Best Arm Identification}
\label{Best Arm Identification}

Several prior algorithms for the problem of best arm identification have been presented over the past few decades, either focusing on the problem from a general perspective or for a specific set of circumstances. Common variations include correlated arm rewards \cite{Gupta2021BestArmII,KAZEROUNI2021365,Hoffman2014OnCA}, identifying the best-$N$ arms \cite{pmlr-v28-bubeck13}, identifying the best arm across multiple overlapping groups \cite{8849327}, and restless arms with evolving states \cite{9965908}. The majority of the these approaches focus on best arm identification for a single bandit, but can also be generalised to a multi-bandit domain.
Out of these past approaches, four algorithms (and variants of them) stand out as being the most applicable to our use case. 

\subsubsection{GapE}
The Gap-based Exploration (GapE) algorithm \cite{NIPS2011_c4851e8e} is one of the first approaches designed specifically for the multi-bandit best arm identification problem. This algorithm favours arms with an expected reward close to the best performing arm (i.e., those with the smallest ``gap''). The authors of this approach also define a variant called GapE-V, which additionally considers the variance in rewards for each arm. 

While on the surface this approach seems well suited for our intended application, the specific characteristics of our general game playing domain highlight some potential issues. Prior observations and experience with these general game frameworks has demonstrated that, while there are certain games where agents frequently achieve different outcomes between repeated trials, a significant portion of games also have a very low variability in agent performance. In these cases, there are several agents that either always win or always lose the game (i.e., reward distributions for game-agent pairs are asymmetric and centred towards the extremities, see Figures \ref{graph:reward_dist_gvgai} and \ref{graph:reward_dist_ludii}). This means that there are often multiple arms for a given bandit with a near-zero gap between them and the best performing arm. 

For our proposed domain, we care primarily about minimising the average simple regret $r(n)$ of our selected arms. As such, if there are several best or near-best arms for a given game, then the decision of which one we select has a negligible impact. GapE appears to have been designed primarily for minimising the average probability of error $e(n)$, and so it instead focuses a large number of arm pulls on these equally high performing arms (i.e., attempts to identify the ``one true'' best arm, rather than being directly concerned with reducing the simple regret of the bandit). This difference often leads to a sub-optimal arm selection strategy from GapE when attempting to minimise the average simple regret across all bandits.

\subsubsection{UCB-E}
The UCB-E algorithm \cite{COLT2010} is a highly exploring variant of the Upper Confidence Bound (UCB) selection policy \cite{10.5555/944919.944941}. One of the most common algorithms for standard (i.e., cumulative reward) multi-arm bandit problems is the UCB1 algorithm \cite{ucb1}, which is able to effectively balance exploration and exploitation. However, best arm identification can be considered a pure exploration problem, as the rewards obtained from each arm during runtime do not affect our performance score (only the final arm pull does). UCB-E attempts to increase the degree of exploration performed by UCB, based on a defined exploration value $a$. If $a = 2\cdot \log(n)$, where $n$ is the current number of rounds, then UCB-E is equivalent to UCB1, with the value for $a$ being increased to incentivise more exploration. 
The effectiveness of UCB-E is highly dependant on the value chosen for $a$, for which the optimal value is often unknown and cannot be easily estimated from past observations. While originally designed for best arm identification in a single bandit case, UCB-E can be generalised to the multi-bandit environment by selecting bandits uniformly and then pulling arms within each bandit using the UCB-E algorithm \cite{NIPS2011_c4851e8e}.

\subsubsection{Successive Rejects}
The Successive Rejects algorithm \cite{COLT2010} repeatedly removes the worst performing arm over multiple rounds of selection until just a single arm remains, a process commonly referred to as ``action elimination'' \cite{6814096}. However, in order for this and other action elimination algorithms to be used to their fullest effectiveness, the total number of arm pulls must be known in advance. This lack of a ``stop anytime'' functionality is a significant downside compared to previous approaches, particularly for our intended use case of general game playing. The amount of time needed to complete a single game trial can be highly variable, meaning that we cannot easily estimate the exact number of trials that can be run within an allotted time. The generalisation of this approach to multiple bandits is also non-optimal, requiring the number of rounds (i.e., arm pulls) to be equally split among all bandits at the start of the process, rather than to bandits with the highest potential benefit during runtime. 

\subsubsection{Sequential Halving}
The original Sequential Halving algorithm \cite{pmlr-v28-karnin13} works in a manner similar to that of the Successive Rejects algorithm, where the worst performing arms of each bandit are removed at the end of each of multiple rounds. However, rather than removing a single arm at the end of each round, half of the remaining arms are removed instead. Sequential Halving also suffers from the same core limitation as Successive Rejects, namely being the requirement that the number of available arm pulls must be known in advance. 

Anytime Sequential Halving \cite{Sagers_2025_AnytimeSH} provides a variant approach that avoids this limitation. This variant performs an initial run of the Sequential Halving algorithm, with a limited budget of arm pulls equal the minimum amount required for a single complete run. Once complete, this ``minimal pull run'' process is repeated, with subsequent reward estimates for each arm incorporating all prior pull results from previous runs. While this Anytime Sequential Halving variant offers the desired stop anytime functionality, it still suffers from being unable to generalise effectively to the multi-bandit domain. Similar to Successive Rejects, arm pulls must be divided evenly across all bandits, resulting in an equal number of trials for each game. 

\subsection{Confidence Intervals}
\label{Confidence Intervals}

Our proposed RCP algorithm (introduced in the next section) relies on a provided confidence interval to estimate the upper and lower reward bounds for a specific arm (based on prior results). While the RCP algorithm can utilise any given upper/lower bound calculation, it should ideally employ a confidence interval that accurately represents the underlying reward distribution. We have therefore chosen to test a variety of suitable confidence interval estimation formulas, evaluating which is most suitable for our intended domain of general game playing.

The majority of confidence intervals described in this section were originally intended for modelling Binomial distributions, where each event has an outcome of either 0 or 1. However, a generalised version of these formulas can still be applied in approximation to non-Bernoulli trials, so long as the outcome is always within the required $[0,1]$ bounds \cite{10.1145/3637528.3671833}. In this case, the proportion of successes ($\hat{p}$) is replaced with the average result of all trials. This generalisation is supported by the Bhatia-Davis inequality, $\sigma^2 \leq \mu (1-\mu)$, which guarantees the true variance is always less than or equal to that assumed by the binomial confidence interval \cite{9a986506-935c-30f3-9482-ebde9bfc2680}. This allows for binomial confidence intervals to be used as a conservative confidence interval estimation in non-Bernoulli cases.

The decision to focus primarily on binomial confidence intervals over more general distribution options is based on the knowledge that the majority of general game outcomes are either a win or a loss. All games used in the GVGAI framework are single-player, meaning that every trial result is either a 0 (loss) or 1 (win). However, the Ludii general game playing system also contains multi-player games, where there is the possibility of a 0.5 outcome for a draw and fractional value outcomes for games with multiple opponents. While this means that the results of Ludii games are not strictly binomial in every case, the fact that the vast majority of game outcomes still end in either a win or loss means that they still exhibit highly binomial distribution patterns in practice (resulting in the need for a generalised binomial distribution interpretation). For alternative problem domains, it may be beneficial to experiment with alternative confidence interval estimations more suited to the underlying result distribution.

\subsubsection{Wilson Score}
\label{Wilson Score}

The Wilson score interval is an approximate binomial proportion confidence interval obtained by inverting a score test for the Bernoulli parameter \cite{afed75f6-c9dc-3279-8854-99fa262b33b1}. Unlike the traditional Wald interval, the Wilson interval avoids relying solely on the asymptotic normality of the sample proportion and instead incorporates an adjustment based on the test statistic. This leads to substantially improved coverage behaviour, particularly when sample sizes are small or when the observed probability of success lies near the extremes (i.e., 0 or 1) \cite{Agresti01051998,doi:10.1080/09296174.2013.799918}. The Wilson score interval for a specified confidence level $(1 - \alpha)$ is defined in formula (\ref{Equation:1}), additionally parameterised by the proportion of successes $\hat{p}$ and sample size $n$.

\begin{equation}
\footnotesize
Wilson(\hat{p},n,\alpha) \equiv (b^-, b^+) = \frac{\hat{p} + \frac{z^2_{\alpha/2}}{2n} \pm z_{\alpha/2} \sqrt{\frac{\hat{p}(1-\hat{p})}{n} + \frac{z^2_{\alpha/2}}{4n^2}}}{1 + \frac{z^2_{\alpha/2}}{n}}
\label{Equation:1}
\end{equation}

\subsubsection{Jeffreys}
\label{Jeffreys Interval}

The Jeffreys interval is a Bayesian binomial proportion confidence interval, derived from the Jeffreys prior for the Bernoulli parameter which corresponds to a $Beta(\frac{1}{2}, \frac{1}{2})$ distribution \cite{1946RSPSA.186..453J}. Similar to the Wilson score interval, Jeffreys interval has been shown to perform well in small sample and unbalanced heavy distributions \cite{10.1214/ss/1009213286}. For a specified confidence level $(1 - \alpha)$ the Jeffreys interval is obtained by taking the lower and upper $\alpha/2$ and $1-\alpha/2$ quantiles of the posterior Beta distribution, see formula (\ref{Equation:2}), additionally parameterised by the number of successes $\hat{x}$ and sample size $n$.

\begin{equation}
\footnotesize
\begin{aligned}
Jeffreys(\hat{x}, n, \alpha) \equiv \Big(
&\, Q_{\alpha/2} \Big(\hat{x}+\frac{1}{2},\, n-\hat{x}+\frac{1}{2}\Big),
\\
&\, Q_{1-\alpha/2} \Big(\hat{x}+\frac{1}{2},\, n-\hat{x}+\frac{1}{2}\Big)
\Big)
\end{aligned}
\label{Equation:2}
\end{equation}

\subsubsection{Clopper–Pearson}
\label{Clopper–Pearson interval}

The Clopper–Pearson interval is an exact binomial proportion confidence interval derived by inverting the cumulative distribution function of the binomial distribution \cite{be7c0fd0-f562-39ad-b8e0-716a276561d1}. Unlike approximate methods, it provides guaranteed frequentist coverage for all parameter values, although this guarantee typically results in intervals that are conservative, particularly when the sample size is small or when the observed proportion is near 0 or 1 \cite{10.1214/ss/1009213286}. For a specified confidence level $(1 - \alpha)$ the lower and upper bounds of the Clopper–Pearson interval are obtained from the $\alpha/2$ and $1-\alpha/2$ quantiles of Beta distributions, see formula (\ref{Equation:3}), additionally parameterised by the number of successes $\hat{x}$ and sample size $n$.

\begin{equation}
\footnotesize
\begin{aligned}
ClopperPearson(\hat{x}, n, \alpha) \equiv \Big(
&\, Q_{\alpha/2} \Big(\hat{x},\, n-\hat{x}+1\Big),
\\
&\, Q_{1-\alpha/2} \Big(\hat{x}+1,\, n-\hat{x}\Big)
\Big)
\end{aligned}
\label{Equation:3}
\end{equation}

\subsubsection{Agresti–Coull}
\label{Agresti–Coull interval}

The Agresti–Coull interval is an approximate binomial proportion confidence interval that improves upon the traditional Wald interval by adding $z_{\alpha/2}^{2}$ pseudo‑observations, split equally between wins and losses, resulting in an adjusted sample size and proportion \cite{Agresti01051998}. This modification yields intervals with markedly better coverage properties, particularly for small sample sizes or when the observed proportion is close to 0 or 1 \cite{10.1214/ss/1009213286}. For a specified confidence level $(1 - \alpha)$, the Agresti–Coull interval is constructed using the adjusted sample size $\tilde{n} = n + z_{\alpha/2}^{2}$ and proportion of successes $\tilde{p} = (\hat{x} + z_{\alpha/2}^{2}/{2})/\tilde{n}$, from which the adjusted confidence interval is computed, see formula (\ref{Equation:4}).

\begin{equation}
\footnotesize
AgrestiCoull(\hat{p},n,\alpha) \equiv (b^-, b^+) = \tilde{p} \pm z_{\alpha/2} \sqrt{\frac{\tilde{p}(1-\tilde{p})}{\tilde{n}}}
\label{Equation:4}
\end{equation}

\subsubsection{Kullback–Leibler Upper Confidence Bound (KL-UCB)}
\label{Kullback–Leibler Upper Confidence Bound}

The KL‑UCB interval constructs confidence bounds for Bernoulli rewards by inverting a constraint on the Kullback–Leibler divergence, rather than using distribution‑free inequalities \cite{pmlr-v19-garivier11a}. Given the proportion of successes $\hat{p}$, the number of arm samples $n$, and the total number of samples across all arms $t$, the upper and lower bounds are defined as the solutions $b^{+}\!\ge\!\hat{p}$ and $b^{-}\!\le\!\hat{p}$ to the implicit equations $n\cdot D_{\mathrm{KL}}(\hat{p}\,\|\,q)\le log(t) + c\cdot log(log(t))$, where $D_{\mathrm{KL}}$ is the Bernoulli KL divergence and $c$ is an exploration constant controlling confidence. Compared with more standard Hoeffding‑type intervals, the KL‑UCB interval yields tighter bounds near the extremes and improved exploration–exploitation trade‑offs \cite{10.1214/13-AOS1119}.

\subsubsection{Probably Approximately Correct (PAC)}
\label{Jeffreys Interval}

The Probably Approximately Correct (PAC) interval is a binomial proportion confidence interval derived from Hoeffding‑style guarantees, ensuring that the empirical success rate remains within a fixed deviation of the true parameter with high probability \cite{10.1145/1968.1972}. In our setting, we adopt an adaptive confidence allocation that spreads the total error budget $\delta$ across all bandits, arms, and number of samples using a $\sum_{t \ge 1}1/{t^2}$ normalisation. Concretely, for an observed proportion of successes $\hat{p}$, arm sample count $n$, number of bandits $M$, number of arms $K$, and total samples $t$, we set:

\begin{equation}
\footnotesize
\alpha_{\mathrm{adapt}}
=
\frac{6\,\delta}{\pi^{2}MKt^2}
\qquad
\beta_{\mathrm{adapt}}
=
\sqrt{\frac{\log\!\left(2 / \alpha_{\mathrm{adapt}}\right)}{2 \max(1, n)}}
\label{Equation:6}
\end{equation}

We then define the confidence interval as:

\begin{equation}
\footnotesize
PAC(\hat{p},n,M,K,t,\delta)
\equiv
\left(
\max\{0,\, \hat{p} - \beta_{\mathrm{adapt}}\},
\;
\min\{1,\, \hat{p} + \beta_{\mathrm{adapt}}\}
\right)
\label{Equation:5}
\end{equation}

This adaptation yields a union‑bound‑controlled guarantee over all bandit arms, with the confidence interval width shrinking at rate $1/\sqrt{n}$, while also growing via the $t^{2}$ weighting. This ensures that the upper/lower confidence bounds for each arm continually increases/decreases (respectively) so long as the arm in question is not pulled. When utilising this confidence interval, our RCP method achieves a fixed-confidence guarantee (proof provided in Appendix \ref{app:pac}).

\section{Regret Change Potential (RCP)}
\label{sec:RCP}

Based on the considerations and restrictions of our general game playing domain, we define the following functionality and design requirements for our proposed approach:

\begin{itemize}
  \item \textbf{Stop Anytime} - We can return the current best identified arm $J_m(t)$ for each bandit $m$ at the end of any round $t$. This allows us set the total number of rounds $n$ to an arbitrarily high value and return the value for $J_m(t)$ whenever needed, or to increase the total number of rounds and continue improving our predictions after the previous best arm predictions $J_m(n)$ have been returned.
  \item \textbf{New Games/Agents} - We can add new bandits or arms to our environment during runtime. This should be possible at the start of a round by initialising any required parameters for each new game-agent pair.
  \item \textbf{Reward Bounds} - Our approach needs to operate within the required [0,1] reward bounds of our domain.
  \item \textbf{Unequal Number of Arms} - Our approach needs to function for sets of bandits with different numbers of available arms $K_m$.
  \item \textbf{Simple Regret} - Our approach should be focused on minimising the average simple regret measure $r(n)$ of our predicted best arms $J_m(n)$.
\end{itemize}

Based on these criteria, we propose our RCP algorithm to tackle the best agent identification problem for general game playing. Algorithm 1 provides the pseudocode for this approach, utilising previously defined variables and terminology, when employing the Wilson Score confidence interval. Similar versions of this algorithm can be created for other confidence intervals, by replacing the $Wilson(\hat{p},n,\alpha)$ call with the desired confidence interval function (e.g., $Jeffreys(\hat{x}, n, \alpha)$ or $AgrestiCoull(\hat{p},n,\alpha)$).

\algrenewcommand\algorithmicrequire{\textbf{Parameters:}}
\algrenewcommand\algorithmicensure{\textbf{Initialise:}}
\begin{algorithm}[t]
\caption{Pseudocode of the RCP algorithm (when using the Wilson Score interval)}\label{alg:ows}
\begin{algorithmic}
\Require number of rounds $n$, alpha value $\alpha$
\Ensure $T_{mk}(0) = 0, \hat{\mu}_{mk}(0) = 0$ for all bandit-arm pairs $(m,k)$
\State Compute $K_{total} = \sum_{m=1}^{M} K_m$
\For{$t = 1,2,...,n$}
    \For{each bandit $m$}
        \State Compute $\hat{k}_{m}^\star = \argmax_{k \in \{1 \dots K_m\}} \hat{\mu}_{mk}(t-1)$
        \State Compute $\hat{\mu}_{m}^\star = \max_{k \in \{1 \dots K_m\}} \hat{\mu}_{mk}(t-1)$
        \For{each arm $k$ of bandit $m$}
        \State Compute $(b^-_{mk}, b^+_{mk}) = Wilson(\hat{\mu}_{mk}(t-1),T_{mk}(t-1),\alpha)$  
        \If{$\hat{\mu}_{mk}(t-1) = \hat{\mu}_{m}^\star$}
            \State $\Delta_{mk} = \hat{\mu}_{m}^\star - b^-_{mk}$
        \Else
            \State $\Delta_{mk} = b^+_{mk} - \hat{\mu}_{m}^\star$
        \EndIf
    \EndFor
    \EndFor
    \State Draw $I(t) \in \argmax_{m,k} \Delta_{mk}$
    \State Observe $X_{I(t)}(T_{I(t)}(t-1)+1) \sim \nu_{I(t)}$
    \State Update $\hat{\mu}_{I(t)}(t) \leftarrow \hat{\mu}_{I(t)}(t-1) + (X_{I(t)}(T_{I(t)}(t-1)+1) - \hat{\mu}_{I(t)}(t-1)) / (T_{I(t)}(t-1) + 1)$
    \State Update $T_{I(t)}(t) \leftarrow T_{I(t)}(t-1) + 1$
\EndFor
\State Return $J_m(n) = \argmax_k \hat{\mu}_{mk}, \ \forall_{m} \in \{1 \dots M\}$
\end{algorithmic}
\end{algorithm}

The general philosophy behind our proposed RCP algorithm is to to select the arm that has the highest potential to impact the overall simple regret across all bandits. To calculate this, the chosen confidence interval (e.g., Wilson Score) is used to estimate the upper and lower confidence bounds $(b^-_{mk}, b^+_{mk})$ of the expected reward of each bandit-arm pair $(m,k)$. 
Note. it is assumed that the chosen confidence interval always returns the maximal range $(0,1)$ when $n$ is zero, meaning that any arm that has not yet been pulled at least once (i.e., $T_{mk} = 0$) is always the preferred selection choice.

The potential regret change $\Delta_{mk}$ for each bandit-arm pair $(m,k)$ is then calculated by comparing its confidence bounds $(b^-_{mk}, b^+_{mk})$ against the expected reward for the current best predicted arm $\hat{\mu}_{m}^\star$. If an arm $k$ is not the current best predicted arm $\hat{k}_m^\star$ for a bandit $m$, then its potential regret change $\Delta_{mk}$ is equal to the difference between its upper confidence bound $b^+_{mk}$ and the expected reward of the best performing arm $\hat{\mu}_{m}^\star$. This represents the potential increase in regret that would be obtained by selecting arm $k$ for bandit $m$, if its true expected reward were equal to its upper confidence bound $b^+_{mk}$. Conversely, if the arm $k$ is the best predicted arm $\hat{k}_m^\star$ for a bandit $m$, then its potential regret change $\Delta_{mk}$ is equal to the difference between the expected reward of the best performing arm $\hat{\mu}_{m}^\star$ (which in this case is identical to its own expected reward $\hat{\mu}_{mk}$) and its lower confidence bound $b^-_{mk}$. This represents the potential decrease in regret that would be obtained by selecting arm $k$ for bandit $m$, if its true expected reward was equal to its lower confidence bound $b^-_{mk}$. 
In summary, we apply an optimistic perspective on all non-best arms by assuming their upper confidence bound is accurate, and a pessimistic perspective on all best arms by assuming their lower confidence bound is accurate. 

The bandit-arm pair with the largest $\Delta_{mk}$ value is pulled to obtain our new reward sample $X_{mk}$, which is then used to update the parameters of our chosen confidence bound (e.g., for the Wilson score interval, this is the arm's expected reward estimate $\hat{\mu}_{mk}$ and pulls counter $T_{mk}$). Once this process has been completed the round is over, and the next round can begin. The current best predicted arm for any bandit $J_m(t)$ can be returned at the end of any round $t$, providing an intermediate measure of algorithm performance. New bandits or arms can also be added at the start of any round by updating the values for $M$ and $K_m$, as well as initialising all values for $T_{mk}(0 \dots t)$ and $\hat{\mu}_{mk}(0 \dots t)$ to 0. Once the specified number of rounds is completed, for each bandit ($\forall_{m} \in \{1 \dots M\}$) we return the arm with the highest expected reward ($J_m(n) = \argmax_k \hat{\mu}_{mk}$).

\subsection{Parameters}

\subsubsection{Number of rounds}
One advantage of our proposed algorithm over alternatives such as Successive Rejects, is that the total number of rounds $n$ has no impact on which arms are pulled each round. This means that the predicted best arm for each bandit $J_m$ can be returned after any number of rounds, without a loss in prediction performance. This ``stop anytime'' nature also allows for alternative stopping criteria, such as algorithm runtime or prediction confidence, to be easily enforced. It is also possible to add additional rounds (i.e., increase the value of $n$) from an arbitrary stopping point, again without any loss in performance. 

\subsubsection{alpha value}
The alpha value directly modifies the upper and lower bounds returned by our chosen confidence interval. A lower alpha value results in a higher degree of "optimism/pessimism" in the potential regret change of each arm, typically leading to an increase in overall arm exploration (vice versa). In line with the typical exploration-exploitation trade-off, higher alpha values often perform better in the short term and lower alpha values tend to perform better in the long term.

\subsection{Uniform Bandit Selection Variant}
\label{uni_bandit}
One potential variant of the RCP algorithm is to select bandits uniformly and then apply the RCP algorithm within each bandit individually. This variation essentially treats our multi-bandit domain as a set of independent best-arm identification problems, where each bandit gets an equal number of pulls. This comes with a potential downside in that we can no longer redistribute arm pulls between bandits (i.e., if we are already confident in the best arm for one bandit, we cannot reallocate it's remaining arm pulls to other bandits). However, it is also possible that this approach could lead to better performance in cases where our RCP approach either over-samples a bandit due to high outcome variance (e.g., stochastic games) or under-samples it due to unlucky outcomes causing a top-performing arm to be dismissed prematurely.

\begin{table*}
\renewcommand{\arraystretch}{1.2}
\caption{Best arm identification algorithms and hyperparameter values to be compared.}
\centering
\begin{tabular}{@{}p{0.18\textwidth}p{0.43\textwidth}p{0.32\textwidth}@{}}
 \toprule
 \textbf{Algorithm} & \textbf{Summary} & \textbf{Hyperparameters} \\
 \midrule
 Random & Random arm selection across all bandits & \\
 Uniform & Equal arm selection across all bandits & \\
 GapE & Selects arms that are close to the current best arm in each bandit & Exploration value ($a$) = [1, 2, 4, 8, 16] \\
 GapE-V & Alternative version of GapE that also considers reward variance & Exploration value ($a$) = [1, 2, 4, 8, 16] \\
 UCB-E & A highly exploring version of UCB that selects arms with the highest upper confidence bound ($a=2\cdot \log(n)$ is equivalent to UCB1) & Exploration value ($a$) = [2, 4, 8, 16] $\cdot$ $\log(n)$, where $n$ = round number\\
 Successive Rejects\textsuperscript{*} & Repeatedly dismisses the worst performing arm in each bandit & \\
 Sequential Halving\textsuperscript{*} & Repeatedly dismisses half the number of arms in each bandit & \\
 Anytime Sequential Halving & Repeatedly performs a shortened version of Sequential Halving, using the minimal number of arm pulls required & \\
 RCP & Selects arms with the highest regret change potential, based on a selected confidence interval
 estimate & Alpha value ($\alpha$) = [0.1, 0.05, 0.01] \newline Confidence interval = [Wilson Score, Jeffreys, Clopper–Pearson, Agresti–Coull, KL-UCB, PAC]\\
 \bottomrule
 \multicolumn{3}{l}{\textsuperscript{*}Requires a fixed total number of arm pulls.} \\
\end{tabular}
\label{table:1}
\end{table*}

\section{Experiments}
\label{sec:Experiments}

\subsection{Game-Agent Datasets}

While the majority of prior research on best arm identification has focused on proving certain theoretical results for specific algorithms, we instead demonstrate the practical application and empirical results of our approach on real-world data. Specifically, we use results obtained from both the GVGAI and Ludii domains across a variety of games and agents.

Results for GVGAI were sourced from the work of \cite{10.1109/CEC48606.2020.9185834}, providing a total of 3,988,557 trials across 108 games and 27 agents. These trials are also split amongst the five distinct levels available for each game, allowing us to include agent results from a subset of levels if desired. To evaluate whether agent performance differences between levels affects our approach's effectiveness, we define both a GVGAI (all levels) dataset that includes trials for each game across all five levels, as well as a GVGAI (level 1) dataset that includes only the 798,168 trials from the first level of each game.

Results for Ludii were obtained by running 29 variants of an Alpha-Beta agent, each using a different heuristic state evaluation function \cite{9619052}. The Alpha-Beta agent was set to Player 1, while any other players were controlled by a standard Upper Confidence bounds applied to Trees (UCT) algorithm \cite{10.1007/11871842_29}. Each agent was given 0.5 seconds of thinking time per move. A maximum turn limit of 500 was applied to all games, and exceeding this value results in a draw for all remaining active players. 
Each agent (i.e., Alpha-Beta heuristic) was run for 100 trials on each of the 1085 games included in Ludii v1.3.9, giving a total of 3,434,627 trials. Any agent heuristics that were incompatible with certain game types were excluded, leading to a varying number of valid agent trials for each game. Results were generated using a single general node (AMD EPYC 7551 @2.55Ghz, 256GB DDR4 @2666Mhz) of the DeepThought high-performance computer \cite{https://doi.org/10.25957/flinders.hpc.deepthought}.

For our presented experiments, rather than running new trials for a chosen game-agent pair we instead sampled rewards randomly (with replacement) from these datasets of previously performed trials. The number of pre-calculated results included in these datasets is large enough to act as a proxy for the true reward distribution of each game-agent pair, and allows for both an increased number of repeated comparison runs and easily reproducible results. Fixing our results datasets also provides us with a defined ``ground truth'' measure of agent performance, which is necessary for the simple regret calculations that are used to evaluate and compare different best-agent identification algorithms. 

In terms of algorithm runtime, most approaches can be run reasonably fast by understanding that the confidence intervals for each game-agent pair are independent of both each other and of the current number of rounds. This means that only the upper and lower confidence bounds of the selected game-agent pair need to be recomputed, and all others will remain unchanged (essentially meaning that just one new calculation needs to be performed each round). The only exceptions to this are the KL-UCB and PAC confidence intervals, which utilise the total number of samples (i.e., current round number) in their calculations. For these confidence intervals, the upper/lower bounds of every game-agent pair must be updated after each round, essentially resulting in $MK$ calculations. However, even for the relatively large number of game-agent pairs in the Ludii dataset, these calculations still take significantly less time than it would take to run a full trial for most games. Given that the total runtime of each algorithm depends heavily on utilised hardware and software optimisations, it has been excluded from the presented results.

All source code, game-agent datasets and performance scores used to produce the presented results is publicly available online.\footnote{\url{https://github.com/stepmat/Best_Agent_Identification_GGP}}

\subsection{Dataset Exploration}

Looking closer at both the GVGAI and Ludii results datasets reveals many of the domain specific factors previously mentioned in Sections \ref{Best Agent Identification for General Game Playing} and \ref{Confidence Intervals}.

\begin{figure}
\centering
\includegraphics[width=1.0\linewidth]{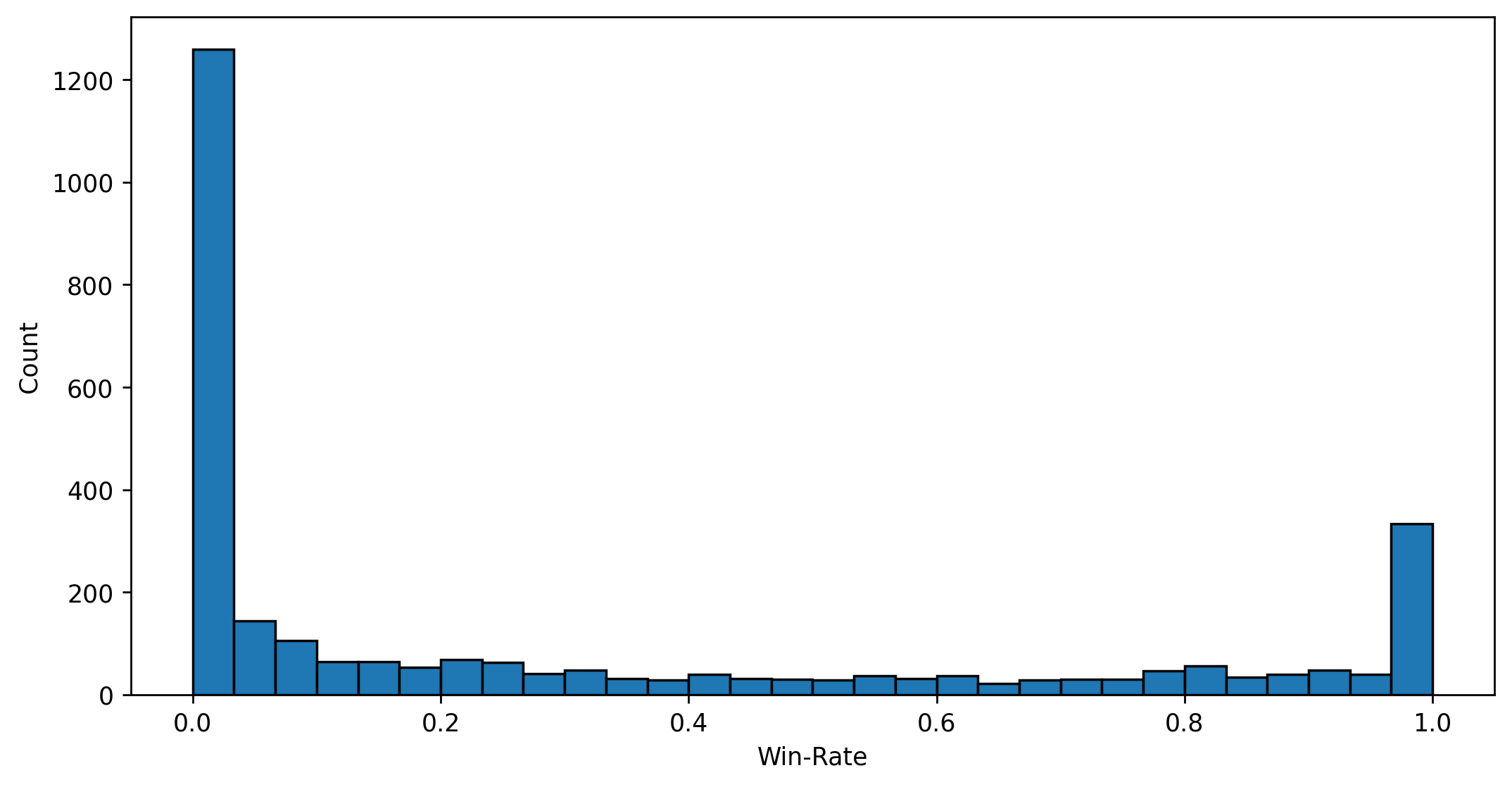}
\caption{Distribution of win-rates for each game-agent pair in the GVGAI (all levels) dataset (total of 2913 game-agent pairs).}
\label{graph:reward_dist_gvgai}
\end{figure}

\begin{figure}
\centering
\includegraphics[width=1.0\linewidth]{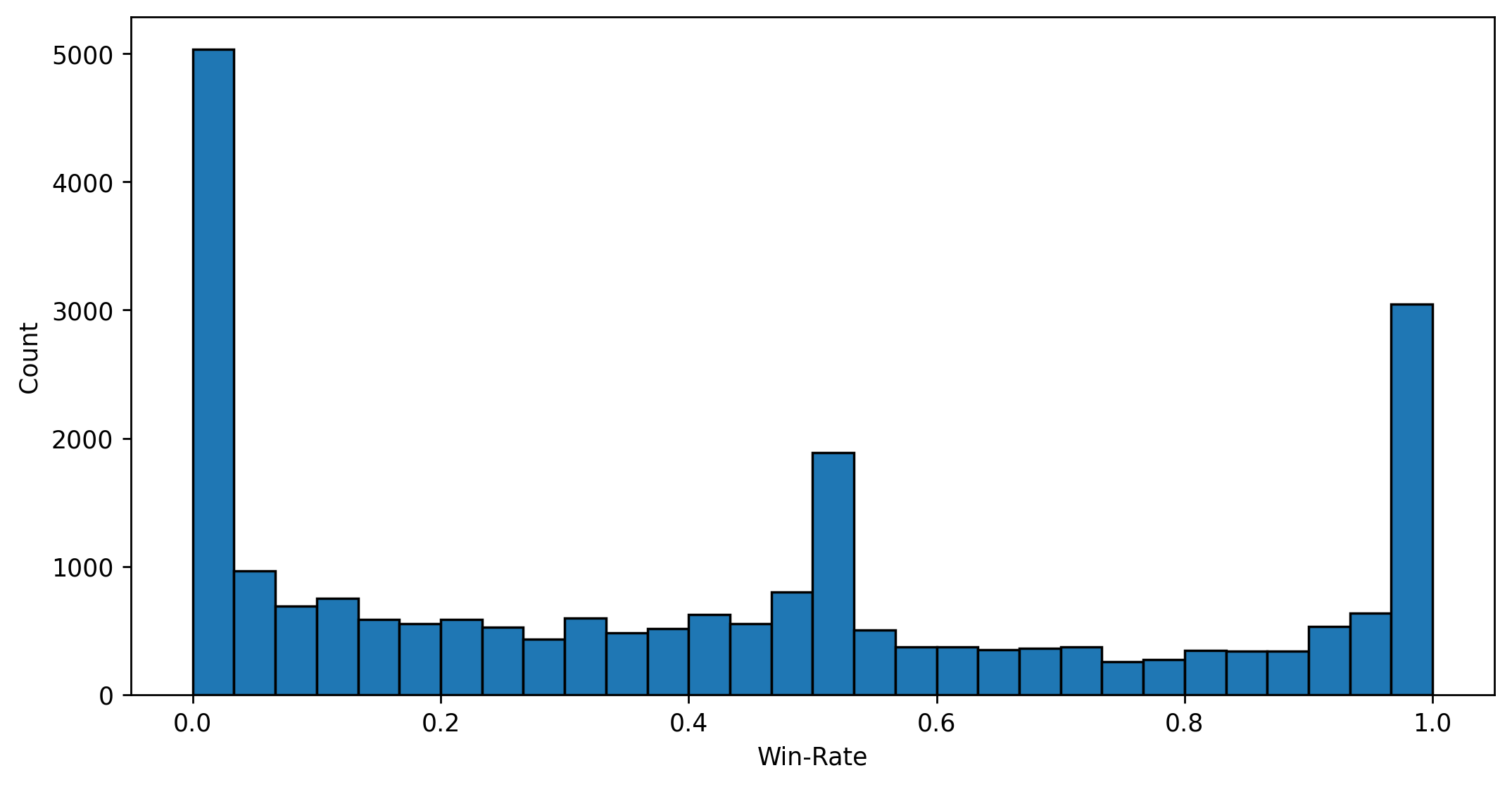}
\caption{Distribution of win-rates for each game-agent pair in the Ludii dataset (total of 23,675 game-agent pairs).}
\label{graph:reward_dist_ludii}
\end{figure}

\begin{itemize}
    \item First, the result distributions for each game-agent pair are often highly skewed towards the 0 and 1 extremities, see Figures \ref{graph:reward_dist_gvgai} and \ref{graph:reward_dist_ludii}, as most agents are reasonably consistent in either winning or losing a game over repeated trials. For the Ludii dataset, we can also observe an additional, albeit smaller, spike around the 0.5 win-rate for games that are frequently drawn. 
    \item Second, while the majority of games have only a single "best agent", there are a significant number of games where multiple agents achieve the same result. For the GVGAI dataset, 26 of the 108 games have multiple top-performing agents ($\sim$24\%), whereas for the Ludii dataset this rises to 370 of the 1085 games ($\sim$34\%). For the purpose of calculating our average probability of error results, selecting any one of these top performing agents would be considered as the optimal choice. 
    \item Lastly, all game trials in the GVGAI dataset have binary outcomes (0 or 1) whereas approximately 22\% of game trials in the Ludii dataset have a non-binary outcome (i.e., a draw or partial win/loss result). This shows that while our Ludii results are not strictly binomial, the majority of game trials ($\sim$78\%) do return either a 0 or 1 outcome.
\end{itemize}

\subsection{Algorithm Comparison}

Using the GVGAI and Ludii game-agent results datasets described above, we evaluate the performance of the multi-bandit best arm identification algorithms described in Table \ref{table:1}. This includes our proposed RCP algorithm, using each of the six confidence intervals described in section \ref{Confidence Intervals}, along with all previous state-of-the-art alternatives described in Section \ref{Best Arm Identification}. To provide a more standard performance baseline, we also evaluate a Random sampling approach that selects a bandit-arm pair at random each time with equal probability (i.e., with replacement) and a Uniform sampling approach that selects each bandit-arm pair an equal number of times (i.e., without replacement).
For algorithms with adjustable hyperparameters, a range of suitable values were tested. Each algorithm and hyperparameter value combination was run on both datasets 10 times, with 50,000 rounds (i.e., arm pulls) in each run. The simple regret of each algorithm was calculated every 1000 pulls, providing us with an average measure of performance across a varying number of rounds.

Given that the optimal initial strategy at the start of any best-arm identification process is to first sample all bandit-arm pairs at least once, we decided to provide a single consistent result from each game-agent pair as an initial win-rate estimate at the start of our experiments. Each best-arm identification approach had its performance estimate for each arm initialised with this first result, as any simple regret measures taken before this time would essentially be identical for all algorithms. The 50,000 arm pull budget provided to each algorithm in these experiments is taken in addition to this initial single pull result for each arm.

\section{Results}
\label{results_section}

\subsection{Hyperparameter Values}

We first discuss the impact that varying the hyperparameter values for each best arm identification algorithm had on it's performance. To avoid cluttering this section with a high number of graphs, the results of this comparison have been placed in the Appendices \ref{app:gvgai_all}, \ref{app:gvgai_0} and \ref{app:ludii}.

\subsubsection{RCP: Alpha Value}

\begin{figure}[!h]
\centering
\includegraphics[width=1.0\linewidth]{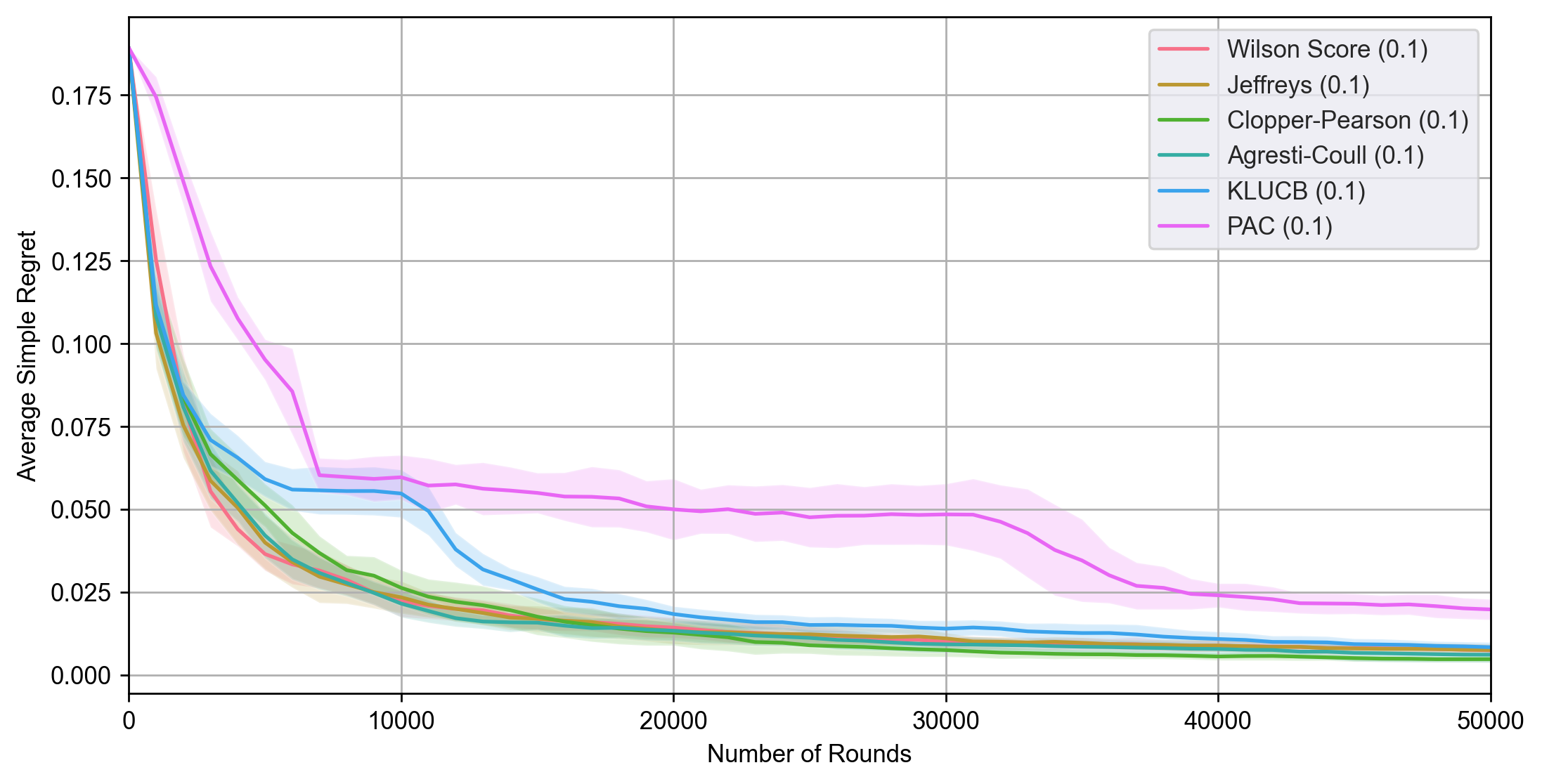}
\caption{Average simple regret for each RCP confidence interval ($\alpha$ = 0.1) when applied to the GVGAI (all levels) game-agent results dataset. }
\label{graph:gvgai_all_rcp_0.1_regret}
\end{figure}

\begin{figure}[!h]
\centering
\includegraphics[width=1.0\linewidth]{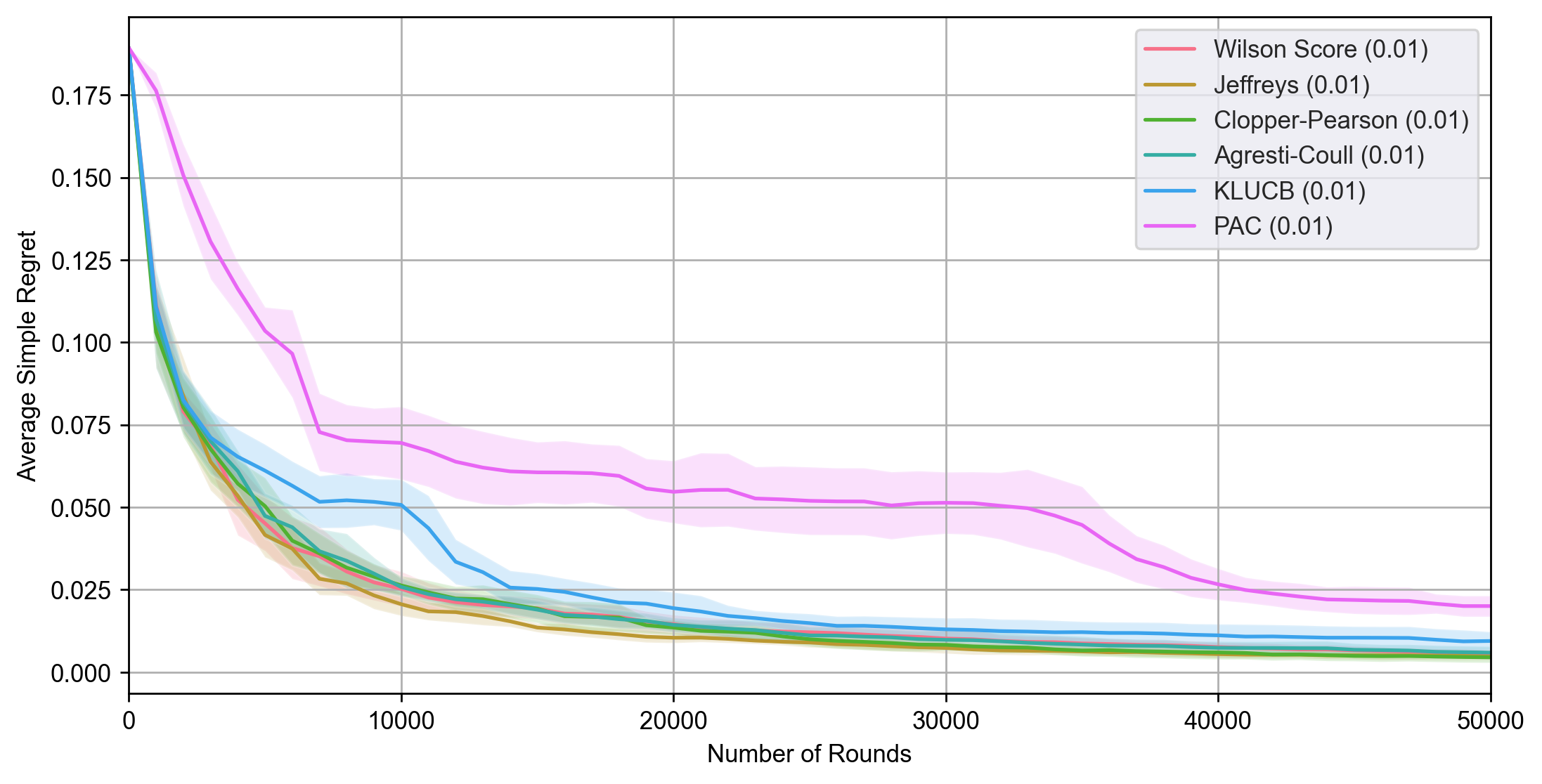}
\caption{Average simple regret for each RCP confidence interval ($\alpha$ = 0.01) when applied to the GVGAI (all levels) game-agent results dataset. }
\label{graph:gvgai_all_rcp_0.01_regret}
\end{figure}

\begin{figure}[!h]
\centering
\includegraphics[width=1.0\linewidth]{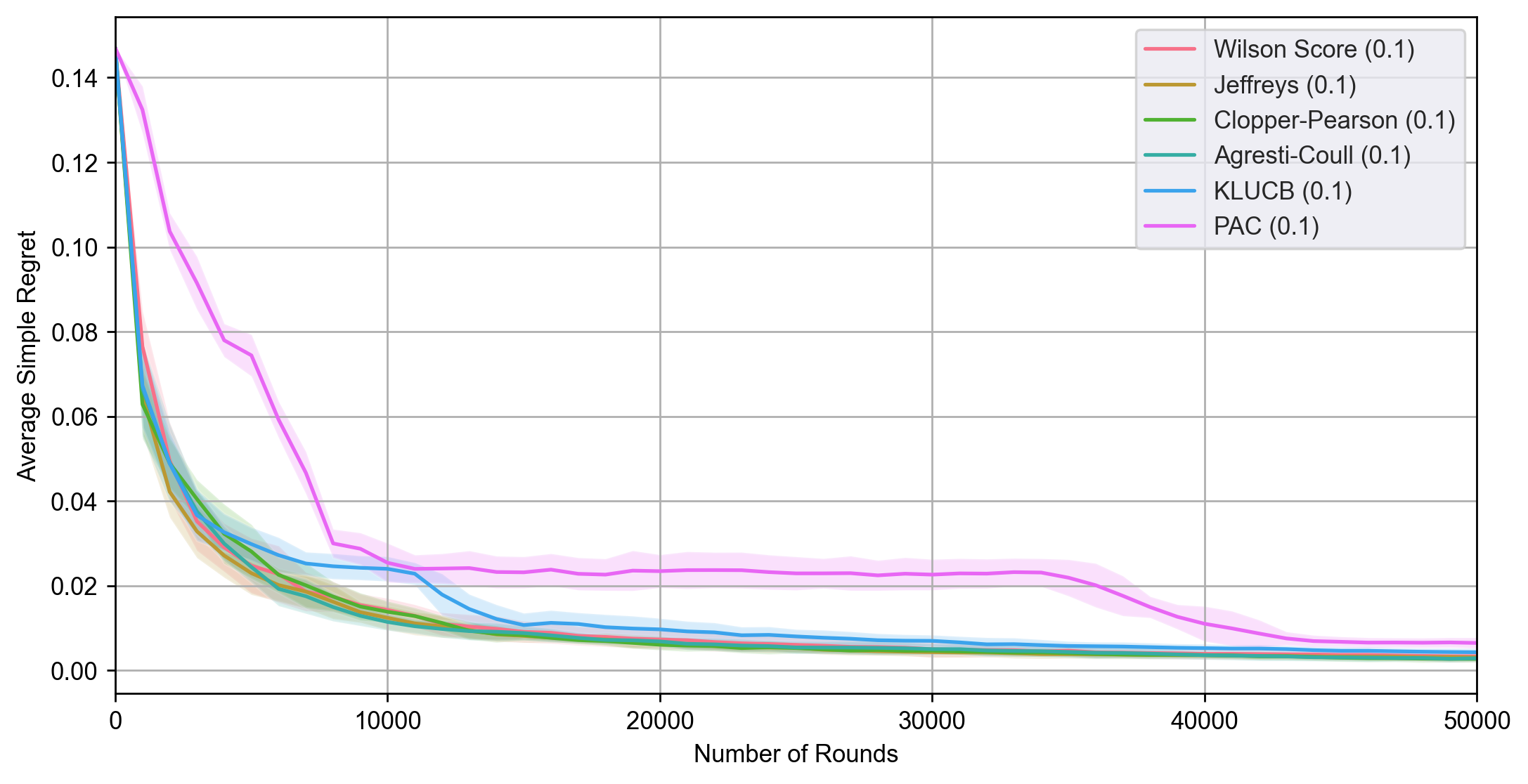}
\caption{Average simple regret for each RCP confidence interval ($\alpha$ = 0.1) when applied to the GVGAI (level 1) game-agent results dataset. }
\label{graph:gvgai_0_rcp_0.1_regret}
\end{figure}

\begin{figure}[!h]
\centering
\includegraphics[width=1.0\linewidth]{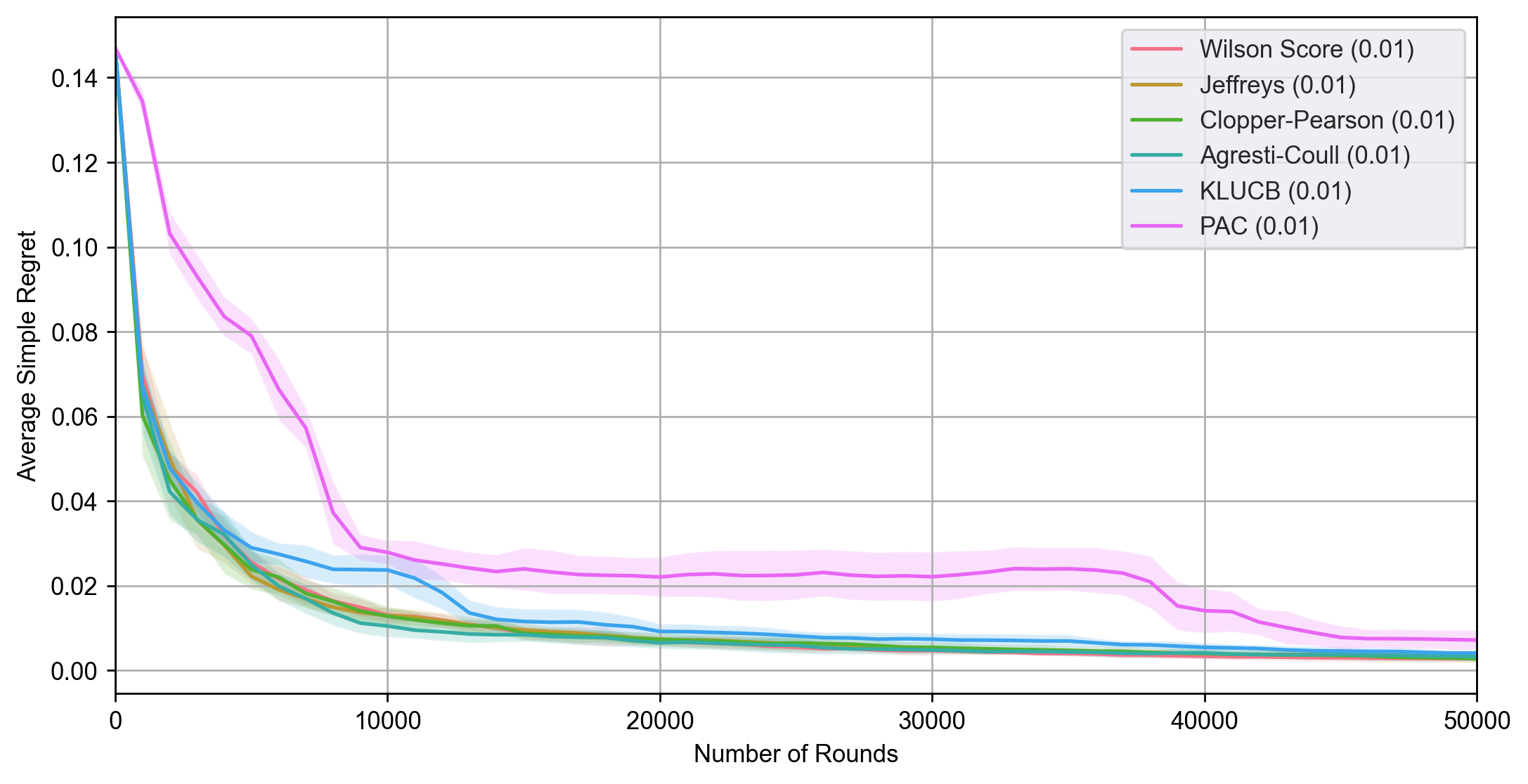}
\caption{Average simple regret for each RCP confidence interval ($\alpha$ = 0.01) when applied to the GVGAI (level 1) game-agent results dataset. }
\label{graph:gvgai_0_rcp_0.01_regret}
\end{figure}

\begin{figure}[!h]
\centering
\includegraphics[width=1.0\linewidth]{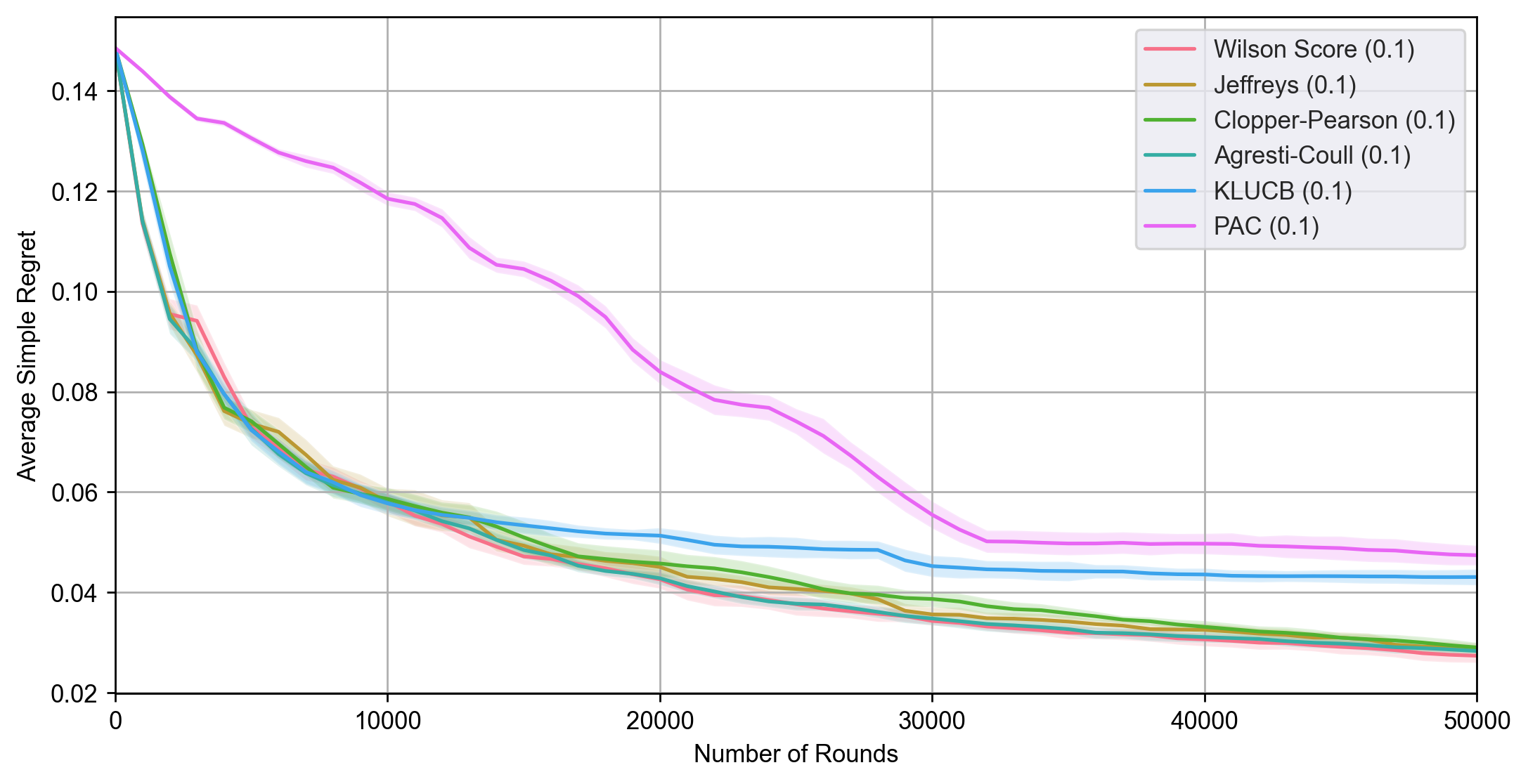}
\caption{Average simple regret for each RCP confidence interval ($\alpha$ = 0.1) when applied to the Ludii game-agent results dataset. }
\label{graph:ludii_rcp_0.1_regret}
\end{figure}

\begin{figure}[!h]
\centering
\includegraphics[width=1.0\linewidth]{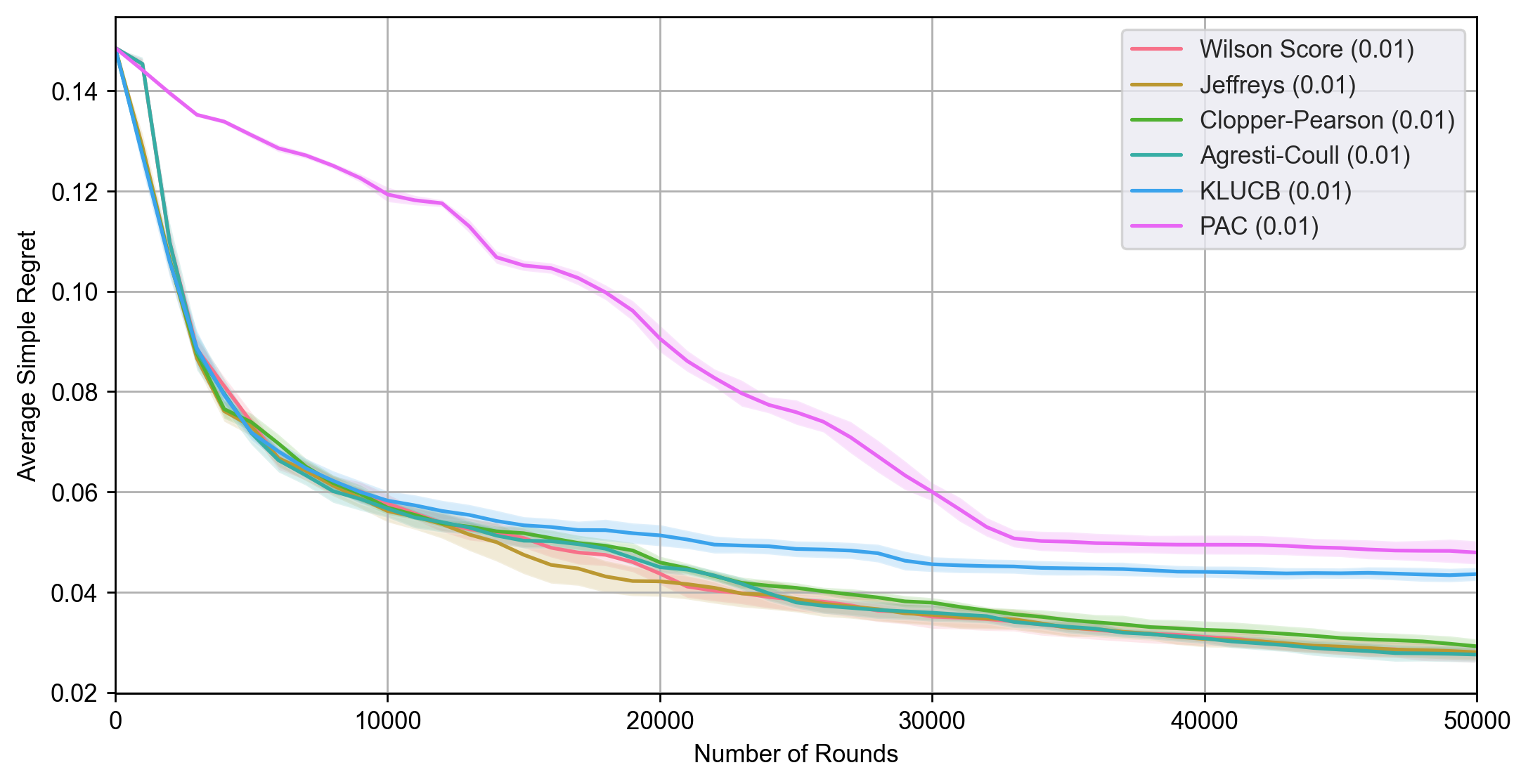}
\caption{Average simple regret for each RCP confidence interval ($\alpha$ = 0.01) when applied to the Ludii game-agent results dataset. }
\label{graph:ludii_rcp_0.01_regret}
\end{figure}

Figures \ref{graph:gvgai_all_rcp_0.1_regret}, \ref{graph:gvgai_all_rcp_0.05_regret} \ref{graph:gvgai_all_rcp_0.01_regret} (GVGAI all levels), \ref{graph:gvgai_0_rcp_0.1_regret}, \ref{graph:gvgai_0_rcp_0.05_regret}, \ref{graph:gvgai_0_rcp_0.01_regret} (GVGAI level 1), and \ref{graph:ludii_rcp_0.1_regret}, \ref{graph:ludii_rcp_0.05_regret}, \ref{graph:ludii_rcp_0.01_regret} (Ludii) show the results of RCP for each confidence interval estimate, using alpha values 0.1, 0.05 and 0.01. From these, we can see that altering the alpha value does not substantially affect the overall trends and performance of these confidence intervals for our RCP algorithm. While there are some minor visible deviations between the alpha values, it is not possible to separate this from normal result variation (i.e., noise) and it would appear that the RCP algorithm performs near equivalently for reasonable alpha values across all of our general game playing domains.

\subsubsection{RCP: Confidence Interval}
\begin{figure}[!h]
\centering
\includegraphics[width=1.0\linewidth]{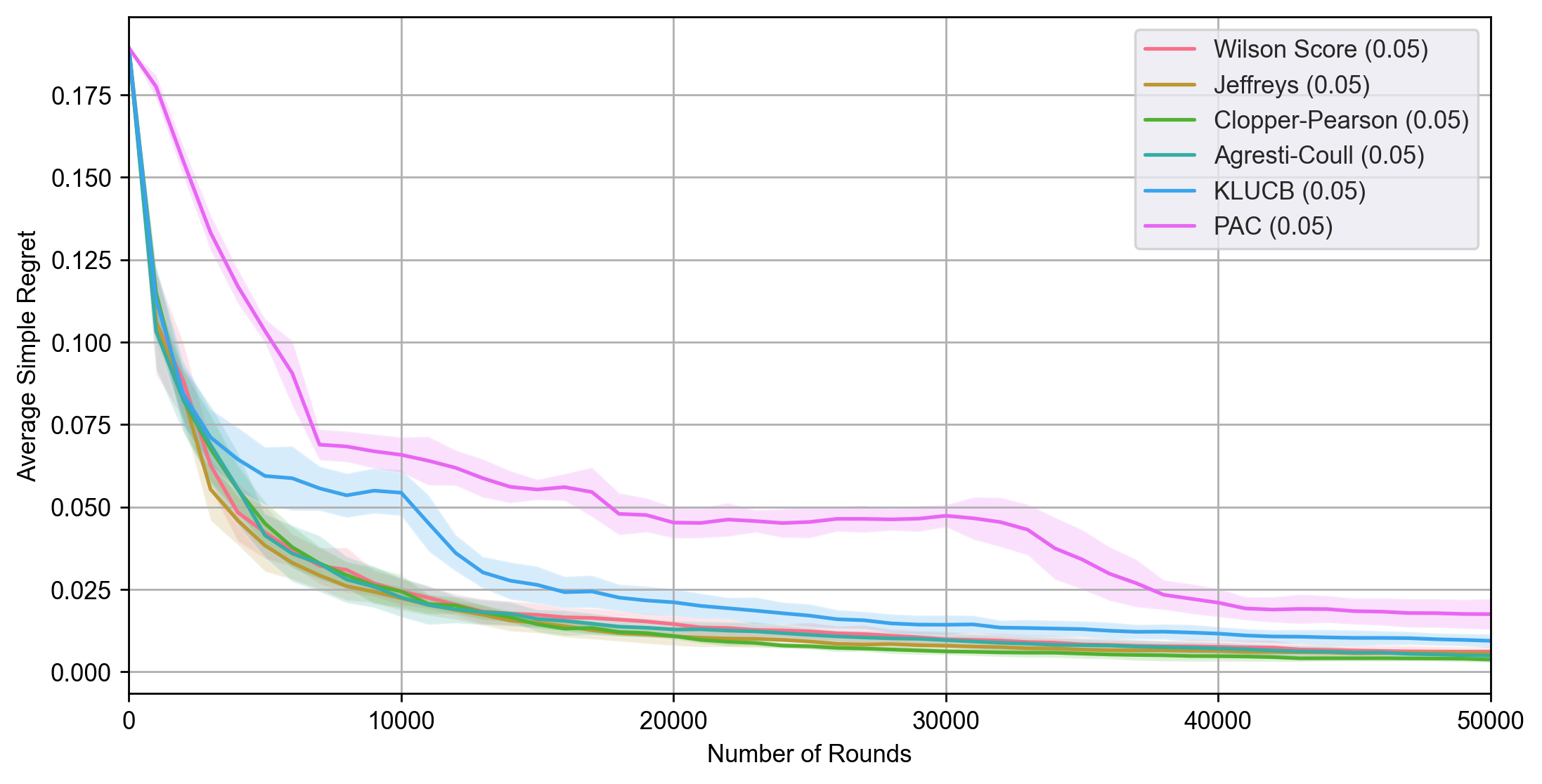}
\caption{Average simple regret for each RCP confidence interval ($\alpha$ = 0.05) when applied to the GVGAI (all levels) game-agent results dataset. }
\label{graph:gvgai_all_rcp_0.05_regret}
\end{figure}

\begin{figure}[!h]
\centering
\includegraphics[width=1.0\linewidth]{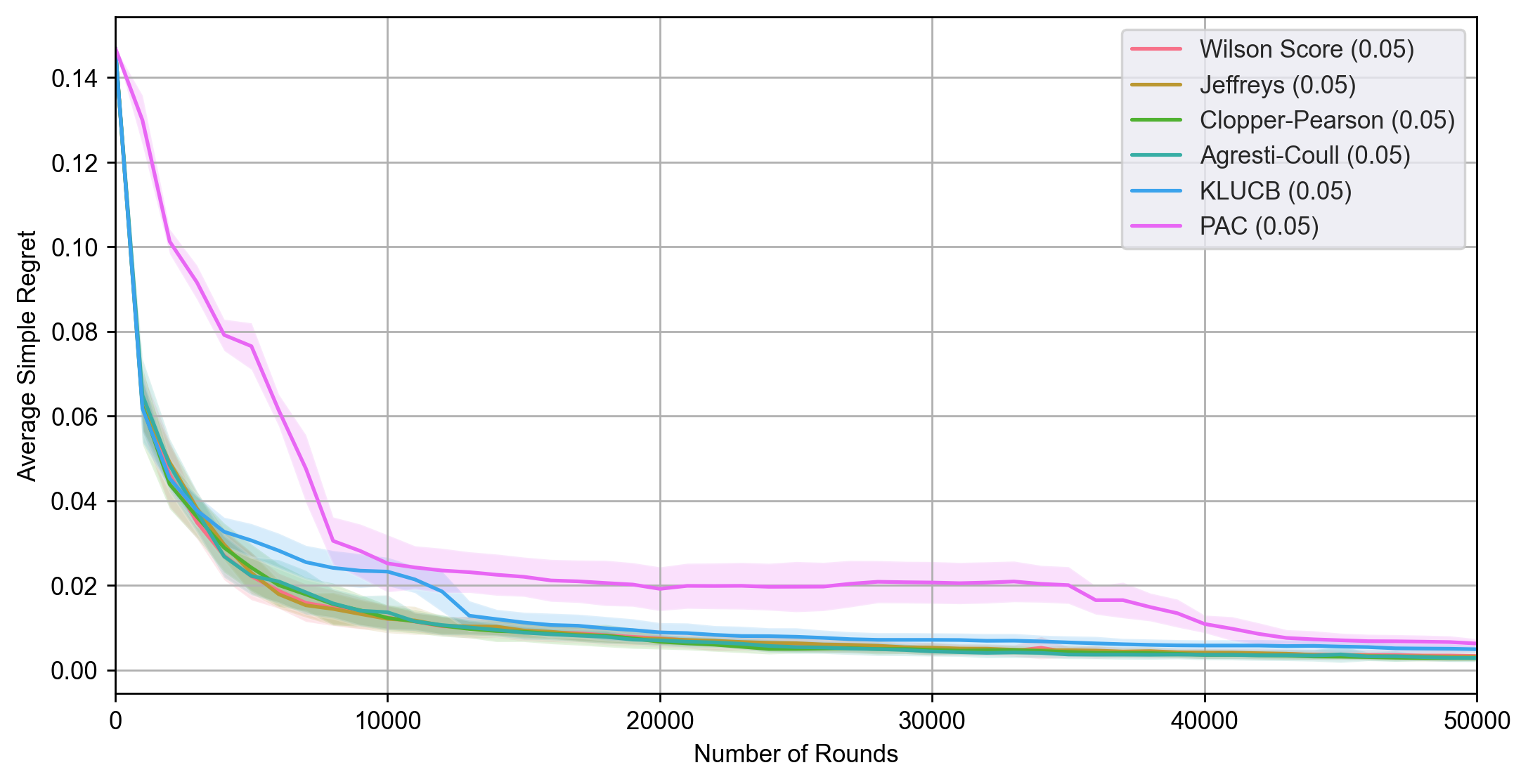}
\caption{Average simple regret for each RCP confidence interval ($\alpha$ = 0.05) when applied to the GVGAI (level 1) game-agent results dataset. }
\label{graph:gvgai_0_rcp_0.05_regret}
\end{figure}

\begin{figure}[!h]
\centering
\includegraphics[width=1.0\linewidth]{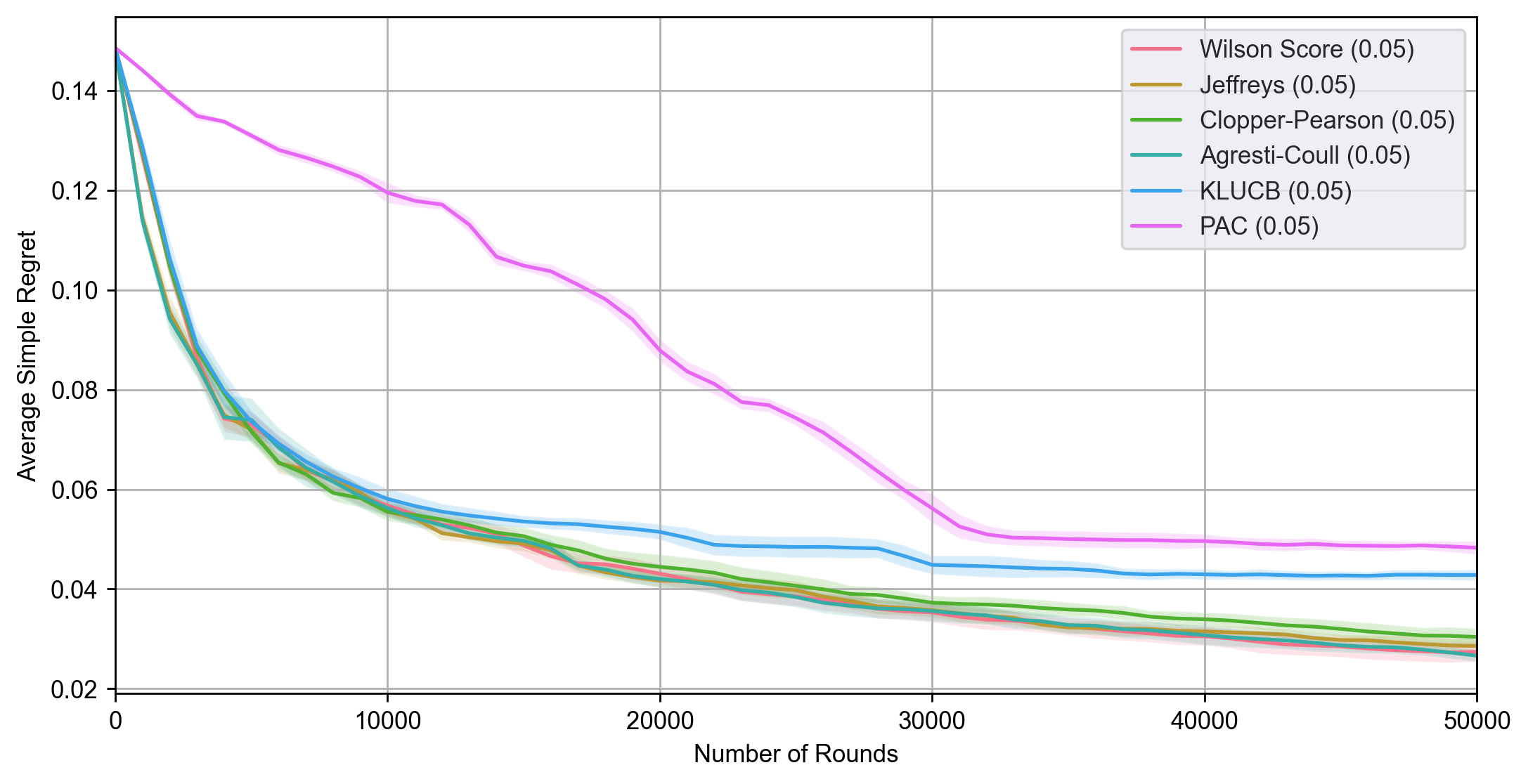}
\caption{Average simple regret for each RCP confidence interval ($\alpha$ = 0.05) when applied to the Ludii game-agent results dataset. }
\label{graph:ludii_rcp_0.05_regret}
\end{figure}

Given that the performance of each RCP confidence interval estimate was largely unaffected by the chosen alpha value, we will focus specifically on the $\alpha$ = 0.05 case (Figures \ref{graph:gvgai_all_rcp_0.05_regret}, \ref{graph:gvgai_0_rcp_0.05_regret} and \ref{graph:ludii_rcp_0.05_regret}). We also see very similar trends across each dataset, and so will discuss the relative performance of each RCP confidence interval estimate collectively across both general game playing domains.

First, we can see that the performance difference in terms of average simple regret for each of the first four confidence intervals (Wilson Score, Jeffreys, Clopper–Pearson and Agresti–Coull) is very minimal. All of these estimates demonstrate near identical trends in terms of regret reduction and variance, indicating that any of these confidence intervals would provide a suitable upper and lower regret bound. 

In contrast, KL-UCB initially performed equally well in the early rounds, but eventually diverged from the other approaches. Interestingly, for the GVGAI datasets (all levels and level 1) KL-UCB showed significantly worse performance between 5,000 and 10,000 rounds, but was able to recover to near equal regret reduction soon after this. For the Ludii dataset, KL-UCB started to diverge from the other top performing estimates after the 10,000 round mark, and this continued to widen as more rounds were carried out.

Lastly, PAC achieved the worst overall performance, achieving a lower average simple regret than the other confidence interval estimates after any number of rounds. Despite this, PAC is the only approach to utilise an adaptive alpha value that ensures widening confidence intervals for each arm as the number of rounds increases (so long as it is not pulled) providing us with a fixed-confidence guarantee. In practice however, it seems that the four binomial proportion confidence intervals implemented (Wilson Score, Jeffreys, Clopper–Pearson and Agresti–Coull) achieve the best regret reduction results up to the 50,000 rounds evaluated.

\subsubsection{GapE / GapE-V / UCB-E: Exploration Value}

\begin{figure}[!h]
\centering
\includegraphics[width=1.0\linewidth]{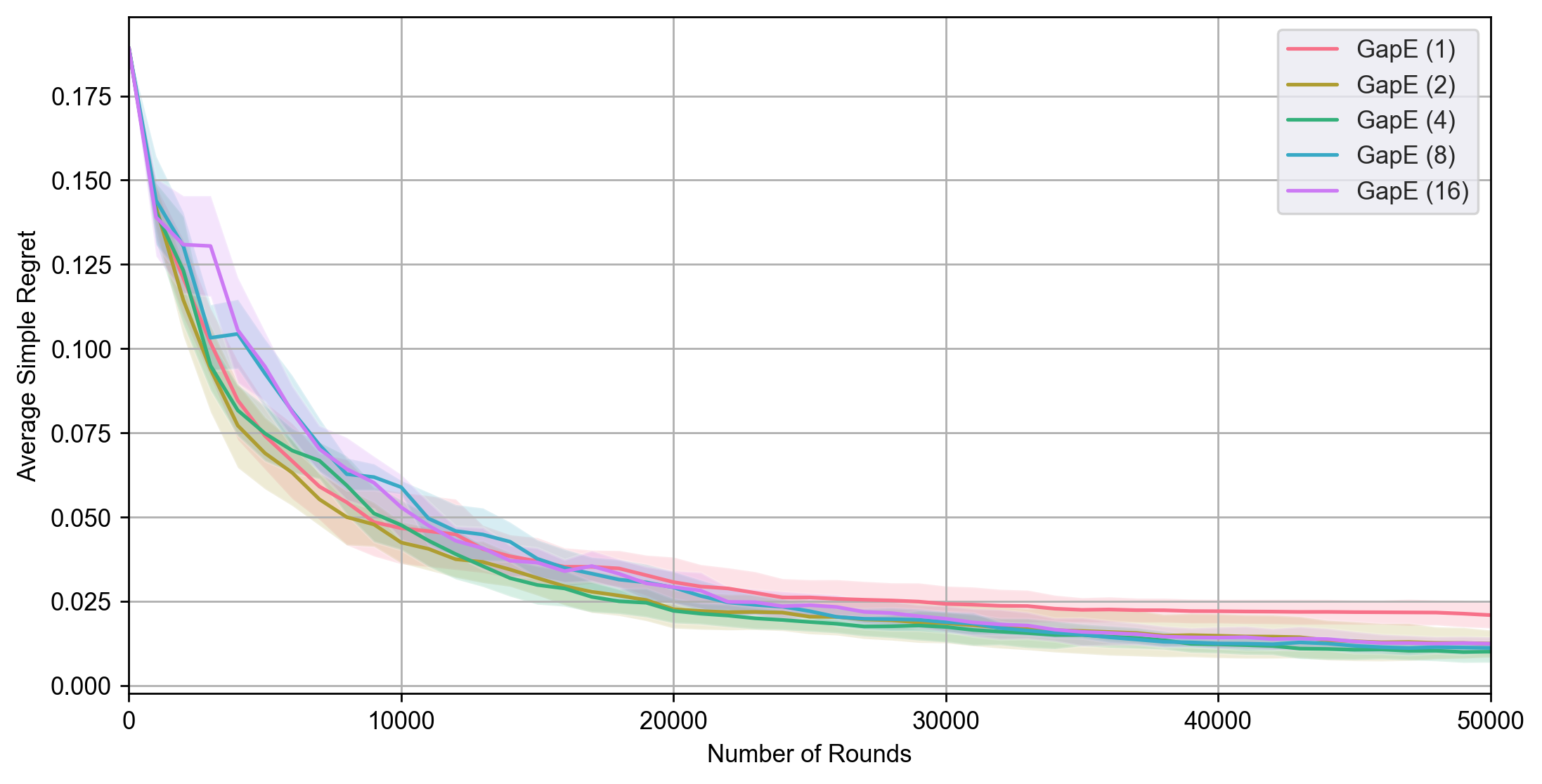}
\caption{Average simple regret for each GapE exploration value when applied to the GVGAI (all levels) game-agent results dataset. }
\label{graph:gvgai_all_gape_regret}
\end{figure}

\begin{figure}[!h]
\centering
\includegraphics[width=1.0\linewidth]{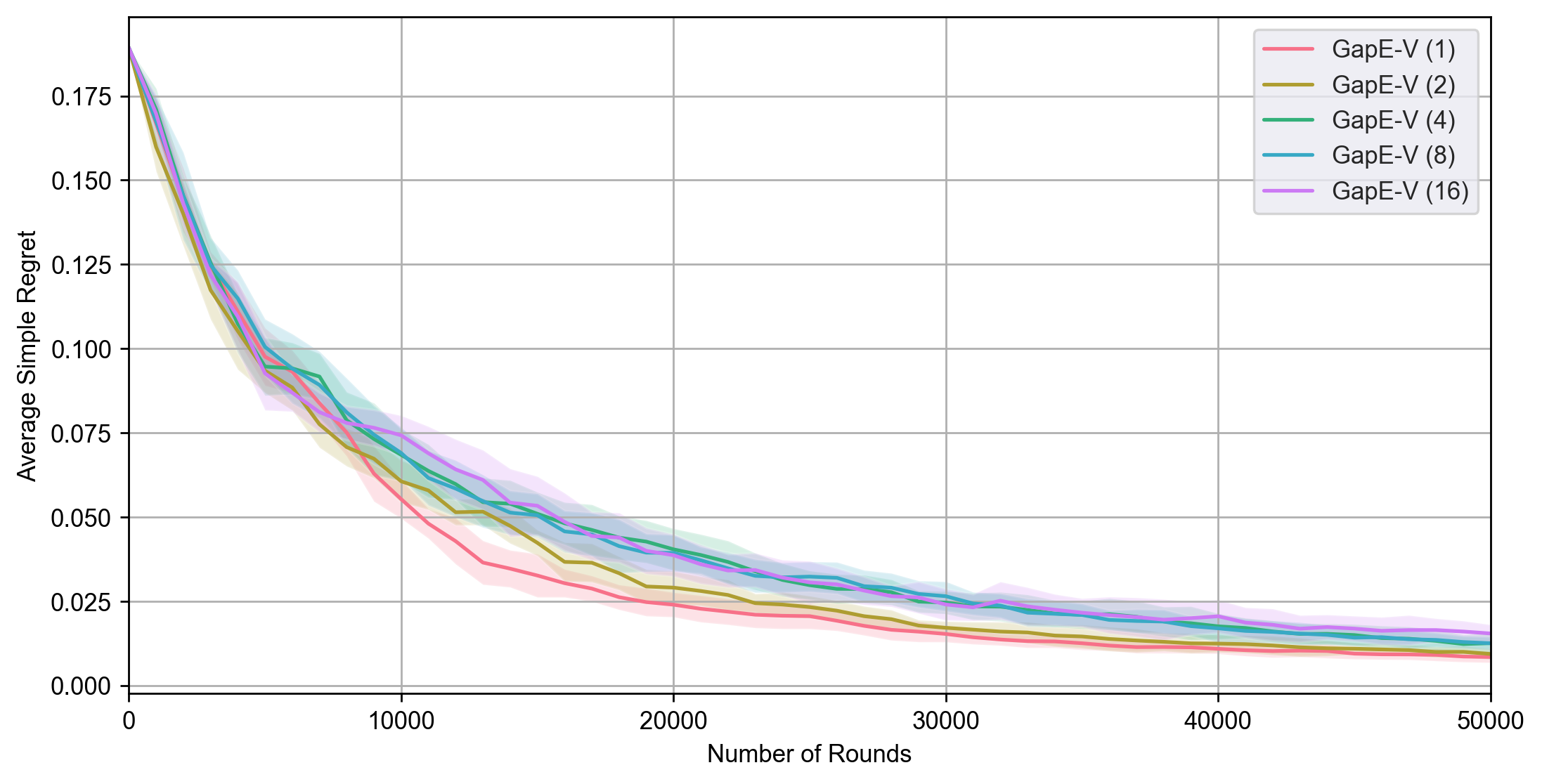}
\caption{Average simple regret for each GapE-V exploration value when applied to the GVGAI (all levels) game-agent results dataset. }
\label{graph:gvgai_all_gapev_regret}
\end{figure}

\begin{figure}[!h]
\centering
\includegraphics[width=1.0\linewidth]{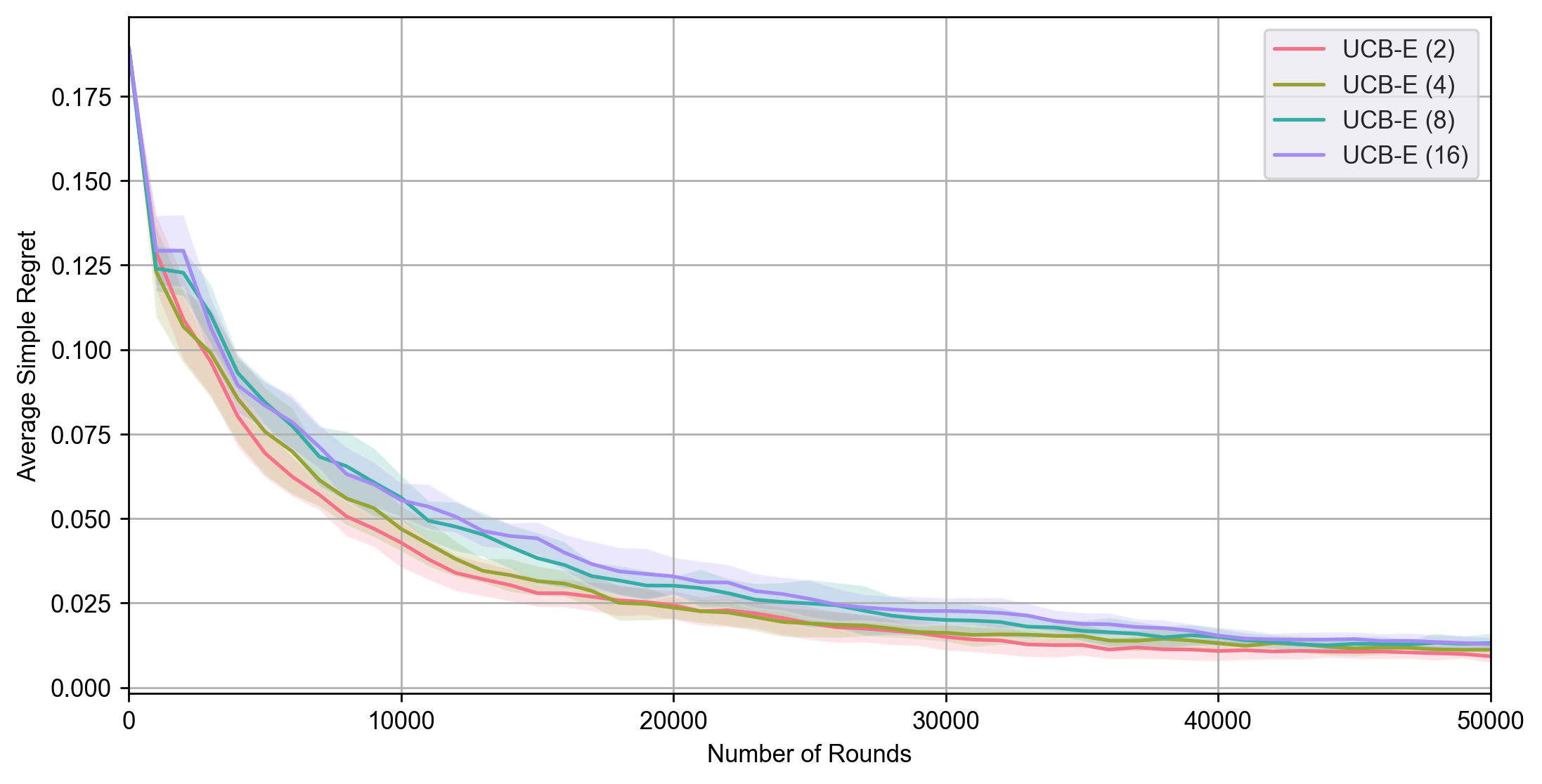}
\caption{Average simple regret for each UCB-E exploration value when applied to the GVGAI (all levels) game-agent results dataset. }
\label{graph:gvgai_all_ucbe_regret}
\end{figure}

\begin{figure}[!h]
\centering
\includegraphics[width=1.0\linewidth]{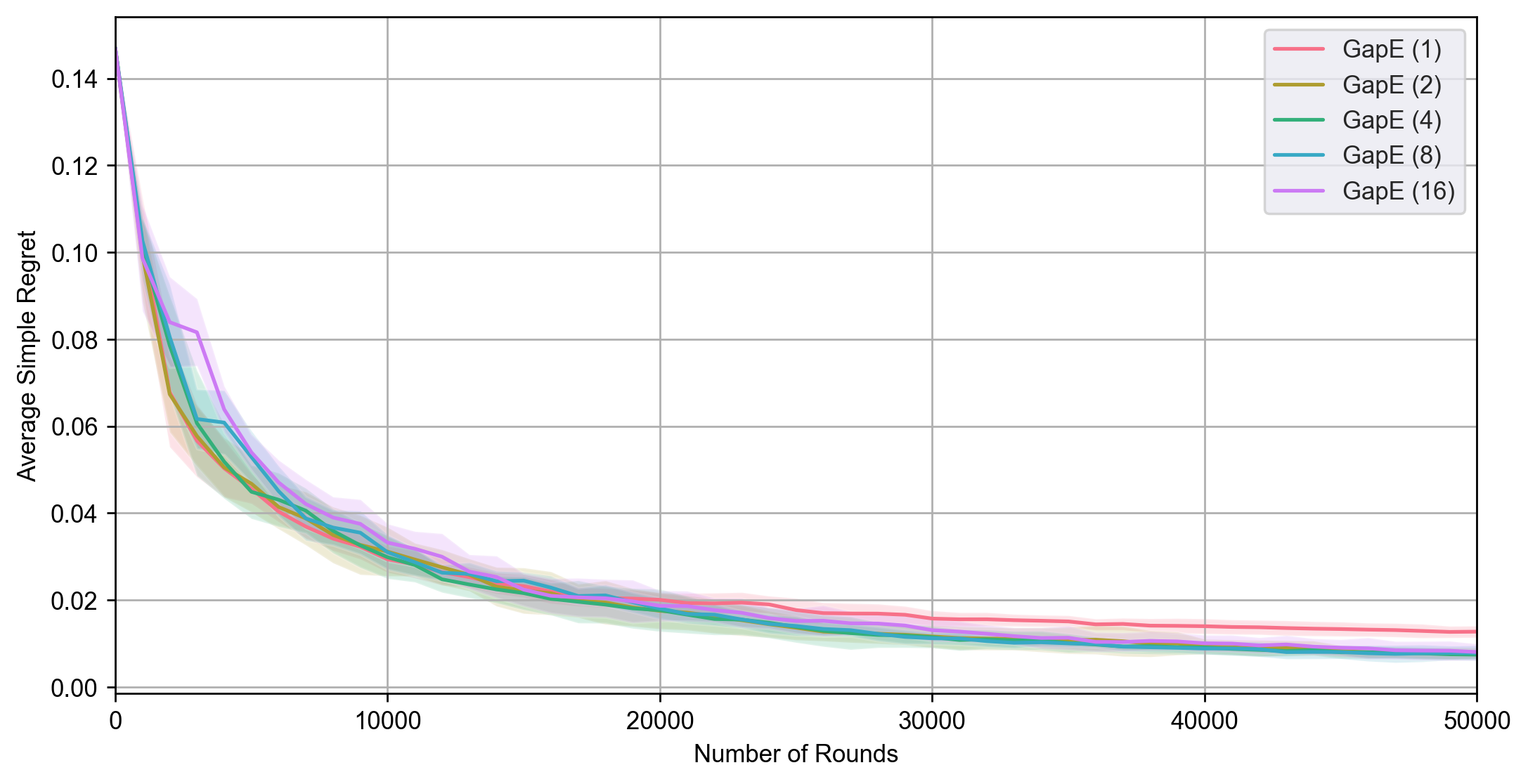}
\caption{Average simple regret for each GapE hyperparameter when applied to the GVGAI (level 1) game-agent results dataset. }
\label{graph:gvgai_0_gape_regret}
\end{figure}

\begin{figure}[!h]
\centering
\includegraphics[width=1.0\linewidth]{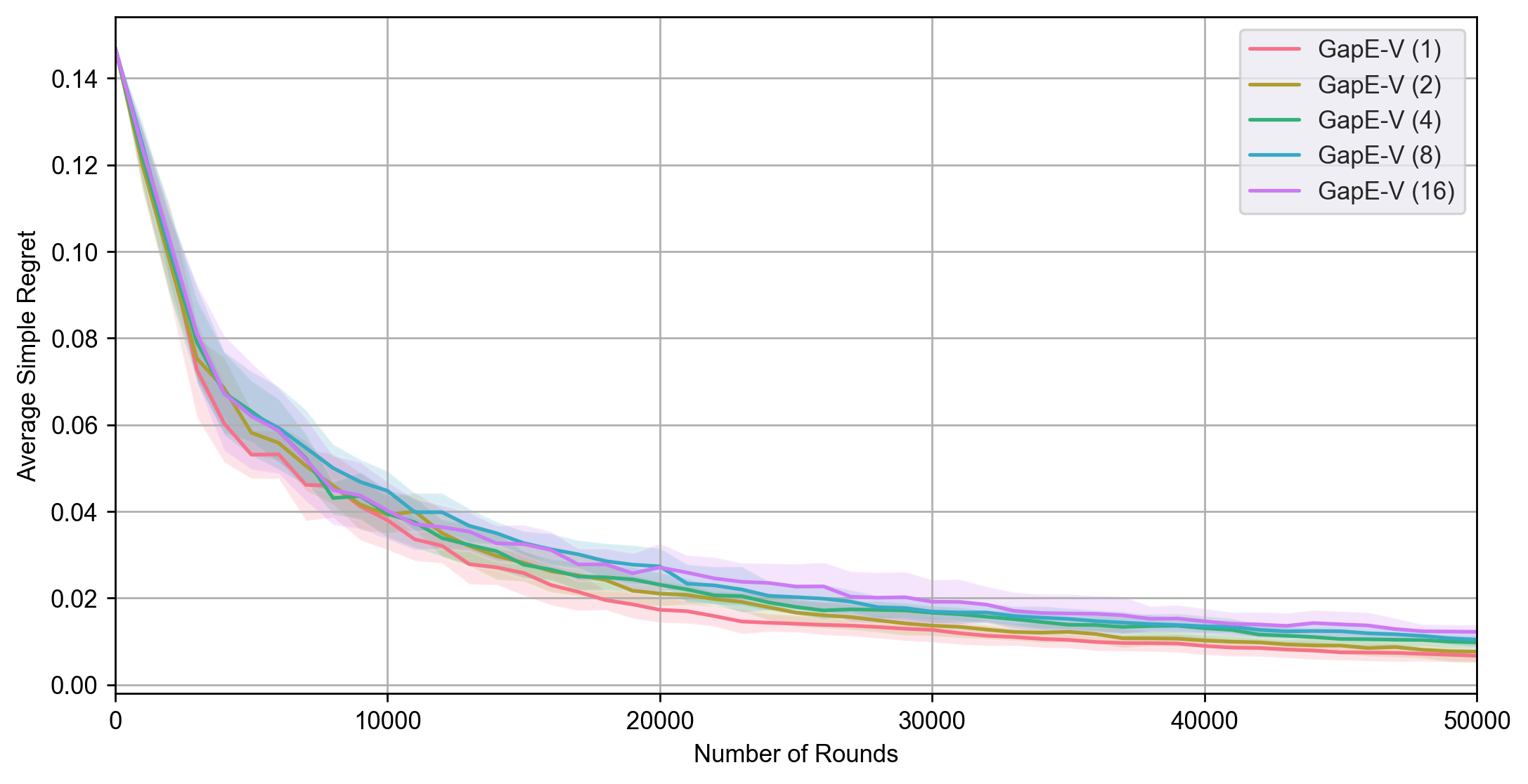}
\caption{Average simple regret for each GapE-V hyperparameter when applied to the GVGAI (level 1) game-agent results dataset. }
\label{graph:gvgai_0_gapev_regret}
\end{figure}

\begin{figure}[!h]
\centering
\includegraphics[width=1.0\linewidth]{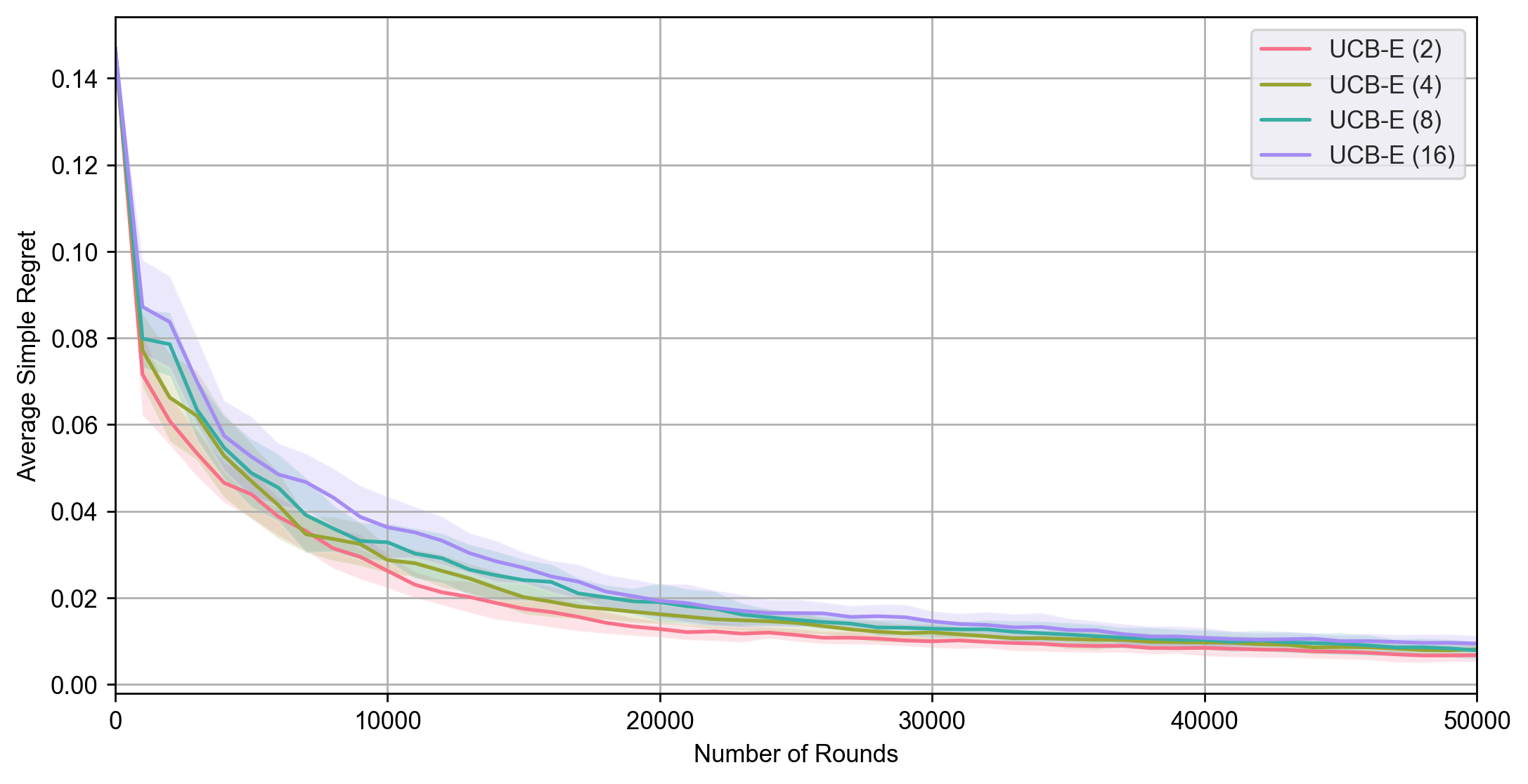}
\caption{Average simple regret for each UCB-E hyperparameter when applied to the GVGAI (level 1) game-agent results dataset. }
\label{graph:gvgai_0_ucbe_regret}
\end{figure}

\begin{figure}[!h]
\centering
\includegraphics[width=1.0\linewidth]{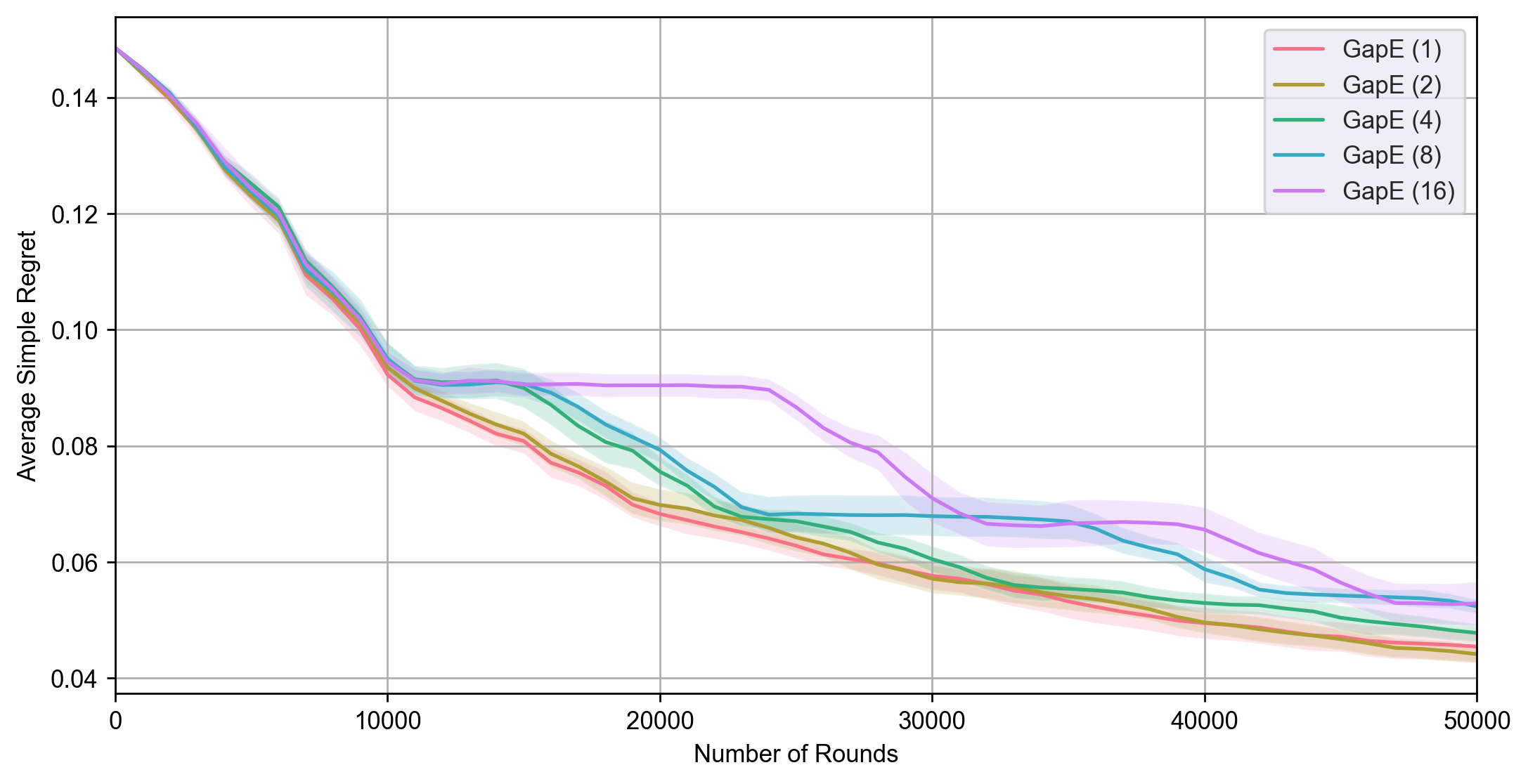}
\caption{Average simple regret for each GapE hyperparameter when applied to the Ludii game-agent results dataset. }
\label{graph:ludii_gape_regret}
\end{figure}

\begin{figure}[!h]
\centering
\includegraphics[width=1.0\linewidth]{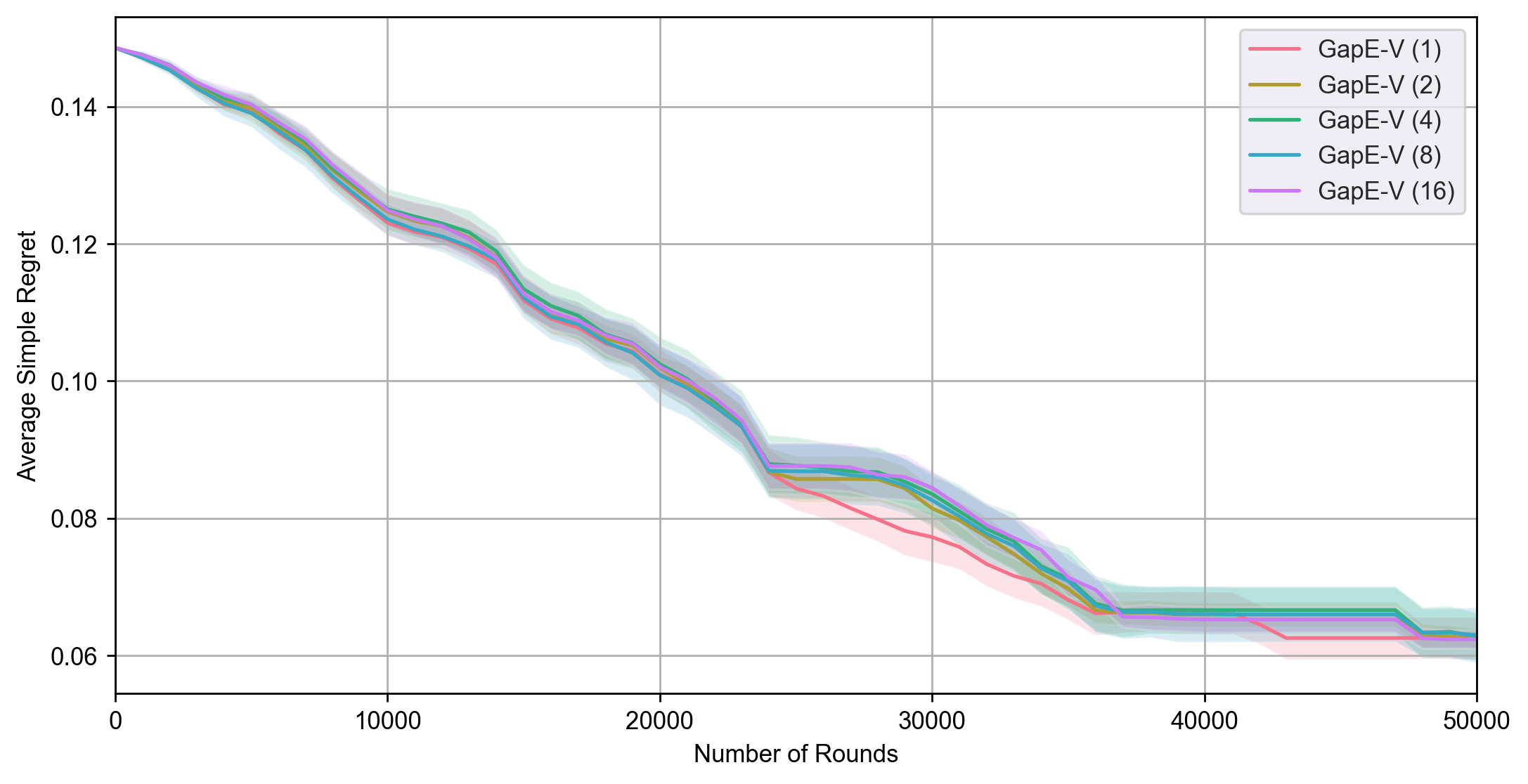}
\caption{Average simple regret for each GapE-V hyperparameter when applied to the Ludii game-agent results dataset. }
\label{graph:ludii_gapev_regret}
\end{figure}

\begin{figure}[!h]
\centering
\includegraphics[width=1.0\linewidth]{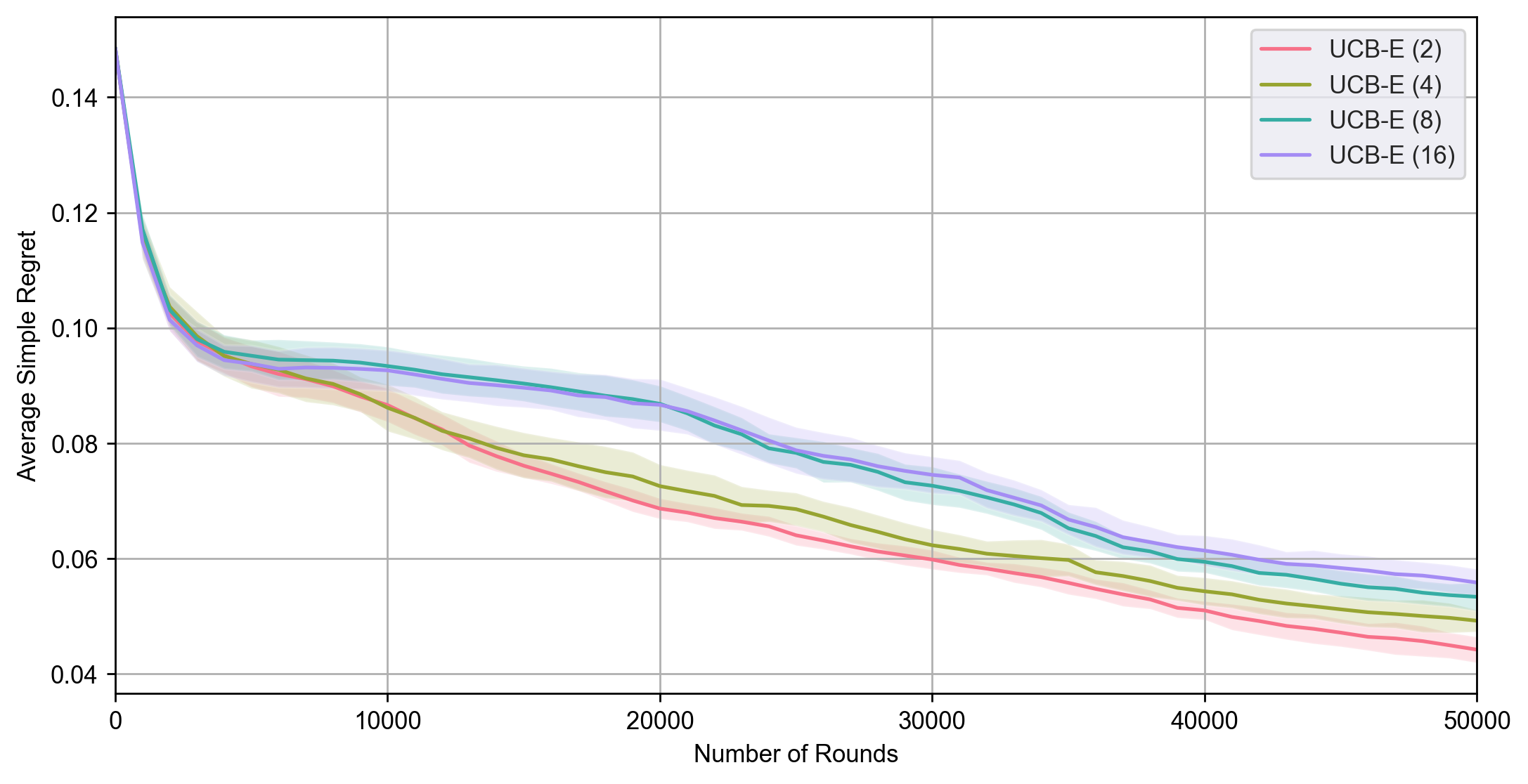}
\caption{Average simple regret for each UCB-E hyperparameter when applied to the Ludii game-agent results dataset. }
\label{graph:ludii_ucbe_regret}
\end{figure}

Figures \ref{graph:gvgai_all_gape_regret}, \ref{graph:gvgai_all_gapev_regret}, \ref{graph:gvgai_all_ucbe_regret} (GVGAI all levels), \ref{graph:gvgai_0_gape_regret}, \ref{graph:gvgai_0_gapev_regret}, \ref{graph:gvgai_0_ucbe_regret} (GVGAI level 1), and \ref{graph:ludii_gape_regret}, \ref{graph:ludii_gapev_regret}, \ref{graph:ludii_ucbe_regret} (Ludii) show results for the GapE, GapE-V and UCB-E algorithms using a range of exploration values. From these, we can see that larger exploration values are typically associated with a higher average regret, particularly after the early rounds. The only exception to this seems to be GapE, where an exploration value of $a$=2 appears to perform slightly better than $a$=1.

\subsubsection{Summary}

Based on these findings, we have selected only the overall best performing hyperparameter value combination for each algorithm to directly compare against each other. More specifically, for GapE ($a$ = 2), for GapE-V ($a$ = 1), for UCB-E ($a$ = 2), and for RCP ($\alpha$ = 0.05, confidence interval = Wilson Score). In practice, we observed that the choice of hyperparameter values for our RCP algorithm had very little impact on its overall performance (with the exception of KL-UCB and PAC confidence intervals). The decision to only compare these specific hyperparameter combinations is simply to prevent graph overcrowding in our presented results. Full results for all hyperparameter values are provided in the Appendix.

\subsection{GVGAI}

\begin{figure}
\centering
\includegraphics[width=1.0\linewidth]{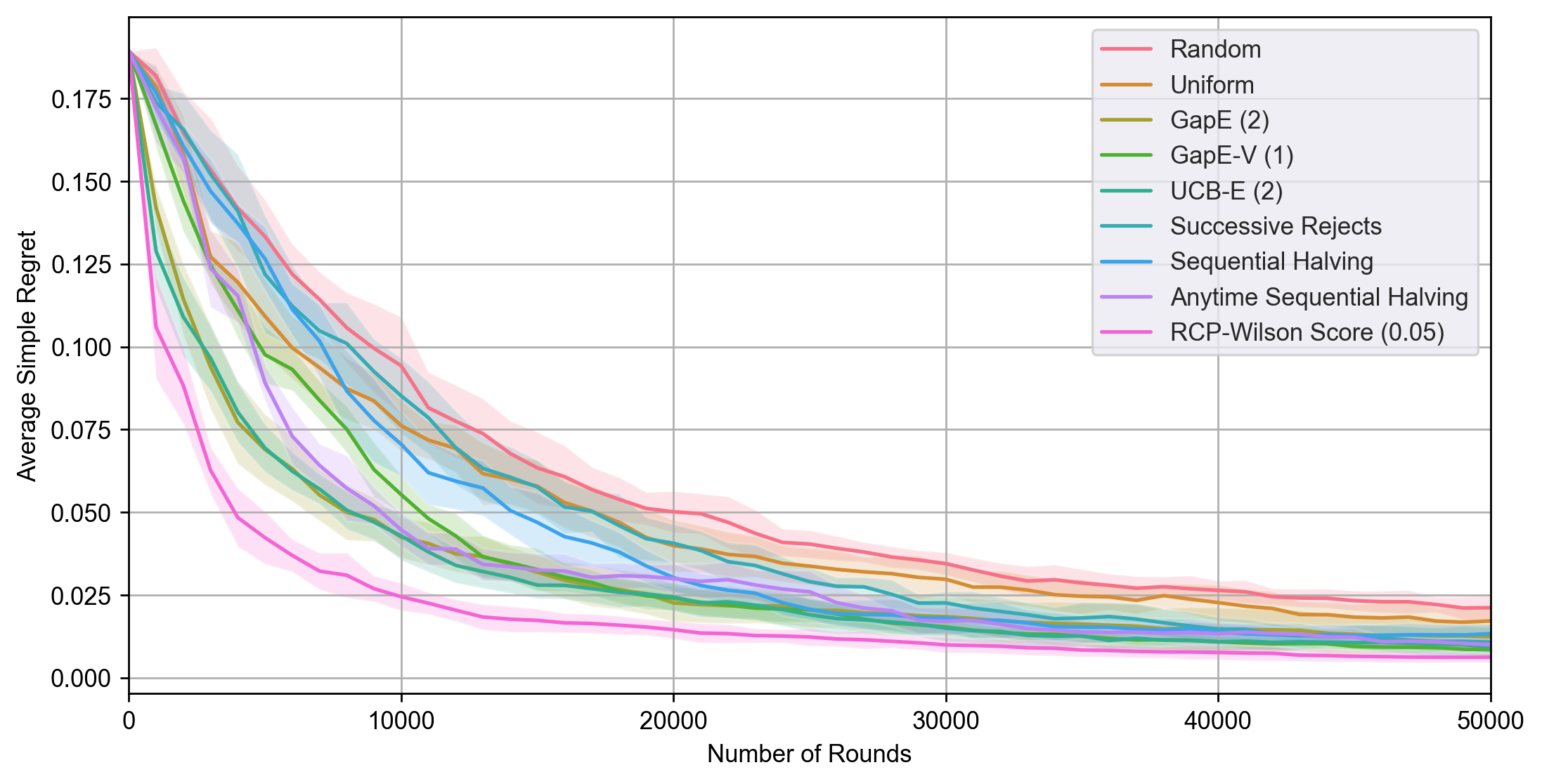}
\caption{Average simple regret for each Best Arm Identification algorithm when applied to the GVGAI (all levels) game-agent results dataset. The shaded uncertainty region indicates plus/minus one standard deviation.}
\label{graph:1}
\end{figure}

\begin{figure}
\centering
\includegraphics[width=1.0\linewidth]{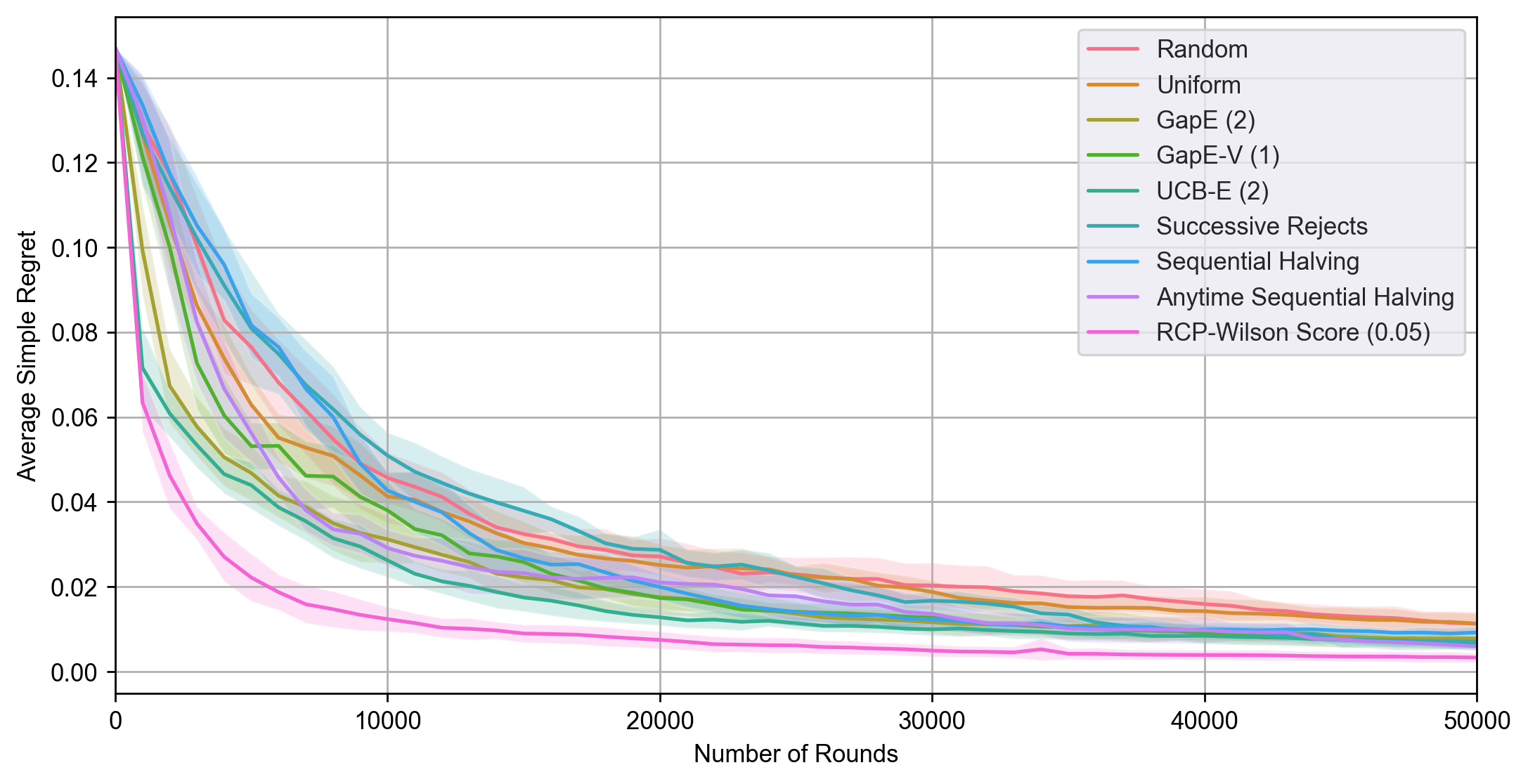}
\caption{Average simple regret for each Best Arm Identification algorithm when applied to the GVGAI (level 1) game-agent results dataset. The shaded uncertainty region indicates plus/minus one standard deviation.}
\label{graph:2}
\end{figure}

The average regret of each best agent identification algorithm for both the GVGAI (all levels) and GVGAI (level 1) datasets are shown in Figures \ref{graph:1} and \ref{graph:2} respectively. From these results we can see that our proposed RCP algorithm provides a substantial performance improvement compared to all previous approaches for any number of rounds. When applied to the GVGAI (all levels) dataset, the simple regret of our RCP approach is approximately 35.7\% smaller than that of the second best approach (UCB-E), and 64.5\% smaller than the more typical uniform sampling approach (averaged across the full 50,000 rounds). The improvement provided by RCP is even more pronounced for the GVGAI (level 1) dataset, with our simple regret being 47.3\% smaller than UCB-E and 70.5\% smaller than uniform sampling over the same 50,000 round period.

Looking closer at these results, we can identify some general trends for each algorithm. Interestingly, the performance comparisons between the GVGAI (all levels) and GVGAI (level 1) datasets are remarkably similar, and so our observations can be jointly applied to both. RCP clearly performs best, achieving a lower average simple regret than all other algorithms for any number of rounds. After this, the second best performing approach is most often UCB-E, with the default UCB1 exploration value of $a = 2\cdot \log(n)$. However, after this the comparison is not as clear cut, with different algorithms performing better after varying round amounts. Gape-E starts off well early on but performs worse as the number of rounds increases, eventually only beating Random, Uniform and Sequential Halving after 50,000 rounds. GapE-V by comparison starts off worse, but overtakes GapE after the 20,000 round mark. Both Successive Rejects and Sequential Halving are designed for a fixed total number of arm pulls, and so initially perform very poorly as the are not intended for early stopping. However, they eventually match the performance of most other approaches once we reach defined 50,000 round end point. The Anytime Sequential Halving algorithm demonstrates its improved ``stop anytime'' functionality over regular Sequential Halving, achieving a much lower average simple regret in the earlier rounds and a near equal performance after 50,000 rounds. Lastly, it is unsurprising that Random and Uniform performed the worst overall, both of which are fairly naive (although still commonly used) evaluation approaches.

\subsection{Ludii}

\begin{figure}
\centering
\includegraphics[width=1.0\linewidth]{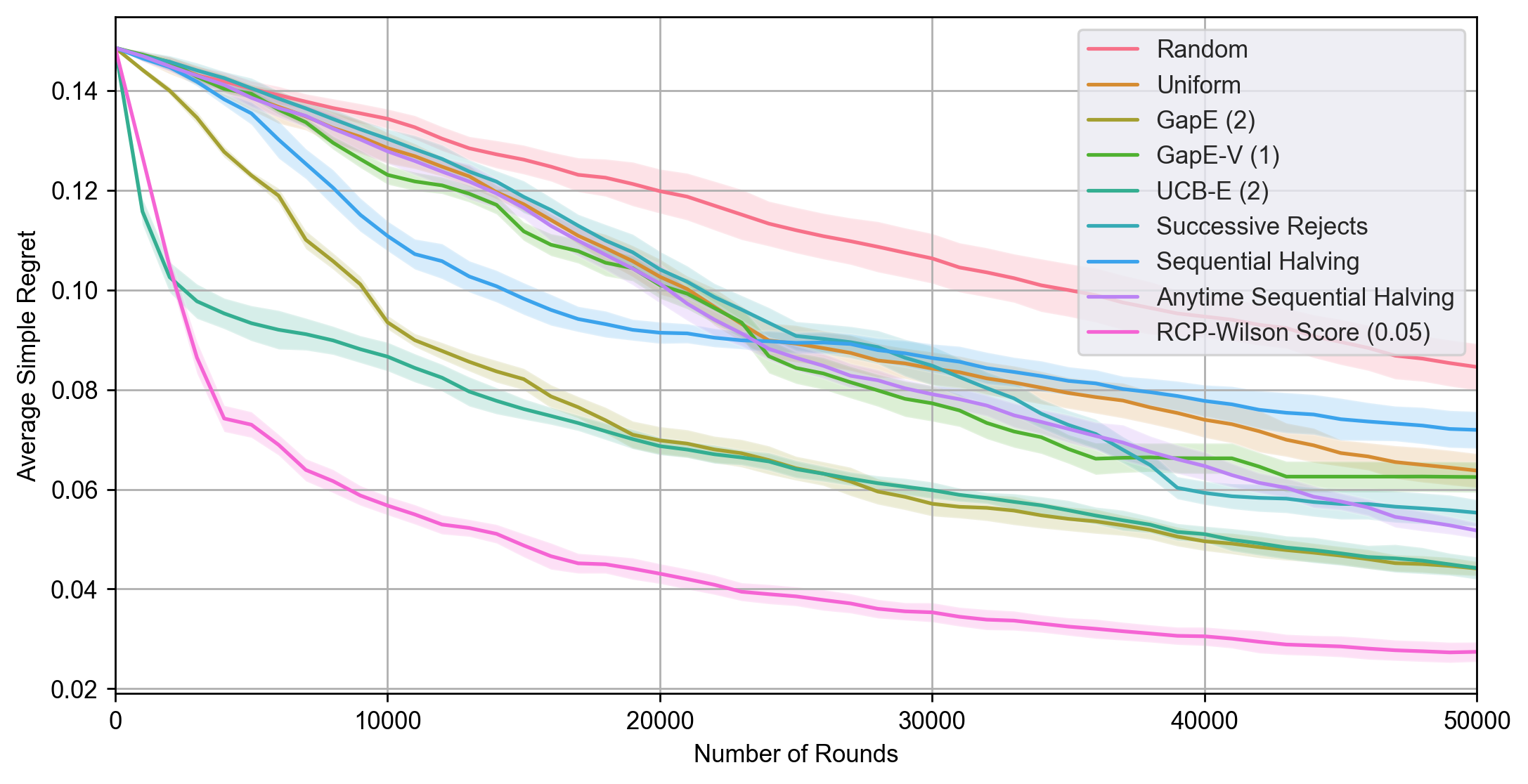}
\caption{Average simple regret for each Best Arm Identification algorithm when applied to the Ludii game-agent results dataset. The shaded uncertainty region indicates plus/minus one standard deviation.}
\label{graph:3}
\end{figure}

The average regret of each best agent identification algorithm for the Ludii game-agent dataset is shown in Figure \ref{graph:3}. Similar to the GVGAI datasets, our proposed RCP algorithm again provides a significant performance increase, achieving a simple regret that is 35.5\% smaller than UCB-E and 55.7\% smaller than uniform sampling when averaged across all 50,000 rounds. 

One immediate observation is that the reduction in average simple regret for the Ludii dataset is less rapid compared to the GVGAI dataset. This could be due to several factors, but is most likely caused by the fact that the Ludii domain simply has more games compared to GVGAI, and hence more bandits to split pulls between. As a result, we would require a much larger number of rounds before our regret would begin to plateau, although we can see this beginning to occur for RCP towards the 50,000 round mark.

In terms of the other other algorithms evaluated, UCB-E started off well (even outperforming RCP for the first few thousand rounds) but quickly worsened and ended up matching the performance of GapE after the full 50,000 rounds. This apparent improvement in GapE compared to the GVGAI results could be due to 50,000 round limit being relatively small compared to the number of Ludii game-agent pairs (23,675) and that with additional rounds the relative performance of GapE may worsen (although this is unconfirmed). GapE-V appeared to perform consistently worse than its GapE counterpart, perhaps due to an insufficient number of rounds for it to get an accurate estimation of each arm's variance. Anytime Sequential Halving initially performed worse than regular Sequential Halving, but did overtake it after 25,000 rounds. Successive Rejects appeared to be doing well until the 40,000 round mark, after which it began to plateau. The results for Successive Rejects and Sequential Halving are particularly odd, flattening well before the 50,000 round limit. The reason for this is largely due to their requirement of a fixed number of arm pulls per bandit, and is discussed in greater detail below. 

\subsubsection{Fixed Round Number Limitations}

One significant drawback of the Sequential Halving and Successive Rejects algorithms is that they often do not use all of the pulls available to them \cite{COLT2010,pmlr-v28-karnin13}. This is because both algorithms distribute pulls across several arm selection rounds ahead of time, requiring that the number of pulls each round be divisible between the remaining arms (which often results in some leftover pulls). This problem is particularly prevalent in situations where the number of available pulls is relatively low compared to the number of bandit arms. Our Ludii dataset presents just such a scenario, with 1085 games and a budget of 50,000 trials meaning that each game is assigned only 46 trials. Given that each game can have up to 29 potential agents that need evaluating, this very small trials-to-agents ratio severely limits the effectiveness of these approaches. Sequential Halving is most impacted by this, with only 26,834 ($\sim$54\%) of its allocated trials being properly used. After this point, any remaining pulls are distributed randomly across all available arms, resulting in a substantially decreased improvement rate compare to previous rounds. Successive Rejects is also affected by this same issue, using just 39,102 ($\sim$78\%) of its available trials before resorting to random pulling, but the fact that it discards its arms one at a time instead of removing half appears to reduce this impact. This issue is also much less prevalent for the GVGAI dataset, likely due to the reduced number of games this dataset has, leading to an increased number of trials per game-agent pair.

\subsection{Average Probability of Error}

\begin{figure}
\centering
\includegraphics[width=1.0\linewidth]{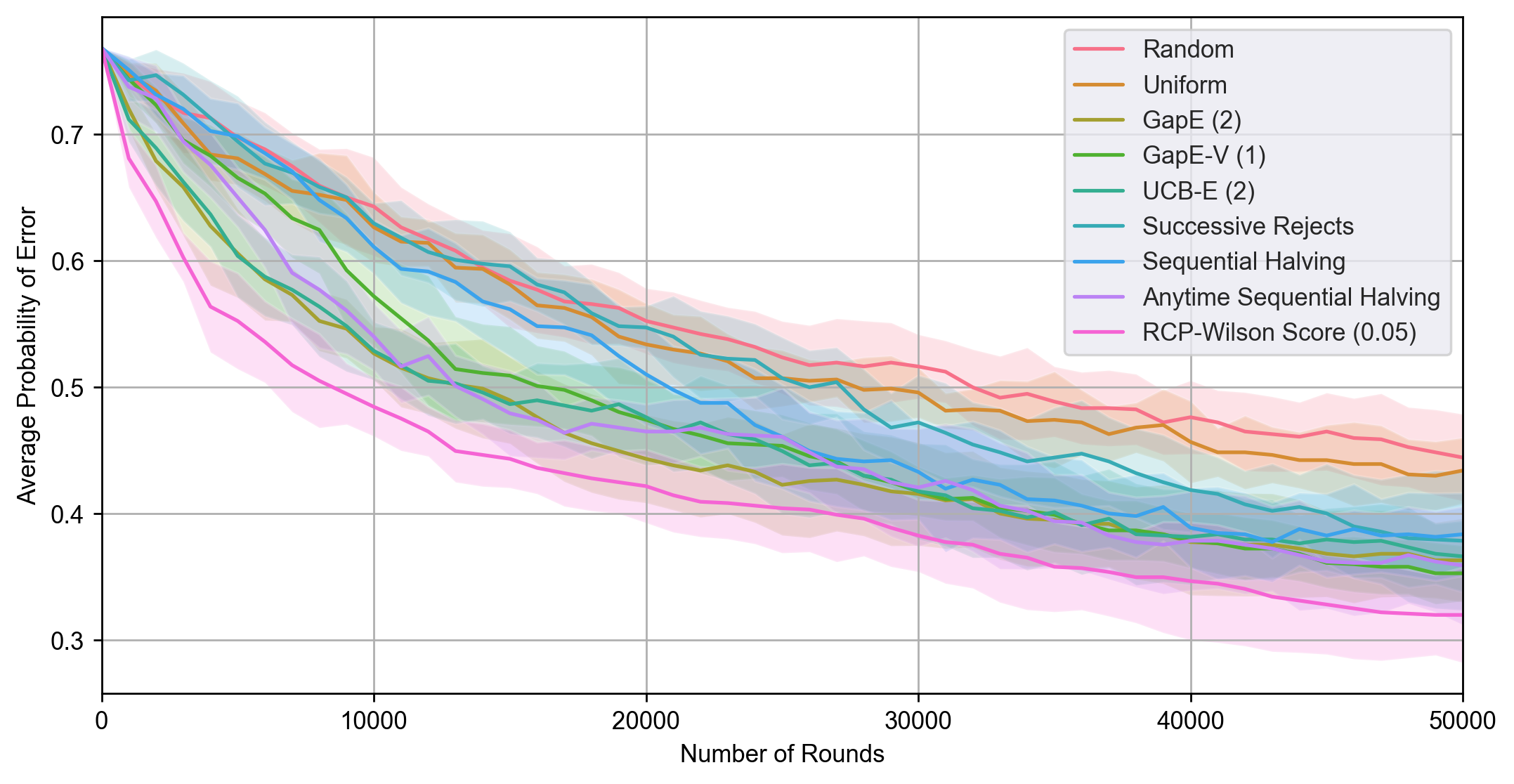}
\caption{Average probability of error for each Best Arm Identification algorithm when applied to the GVGAI (all levels) game-agent results dataset. }
\label{graph:a1}
\end{figure}

\begin{figure}
\centering
\includegraphics[width=1.0\linewidth]{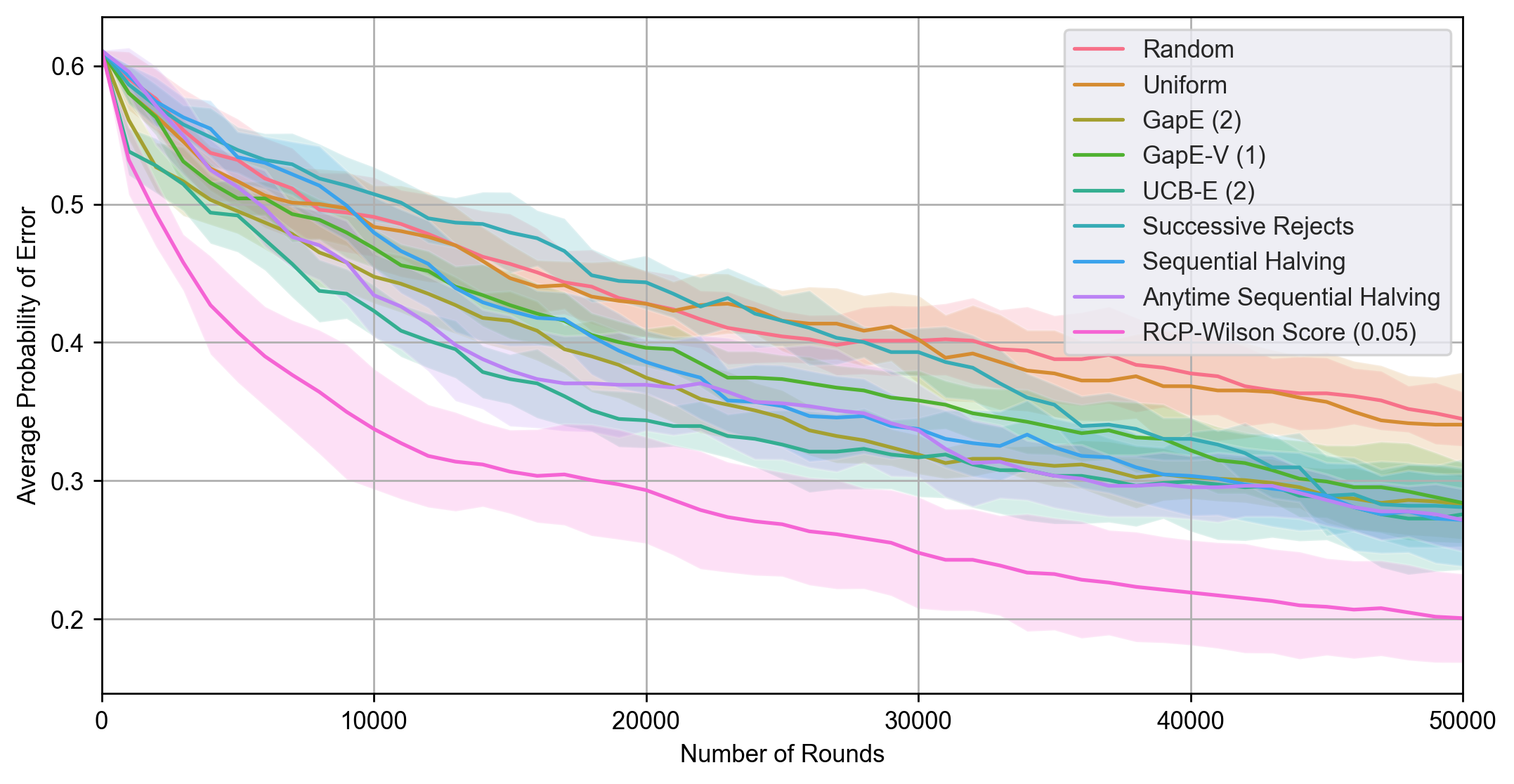}
\caption{Average probability of error for each Best Arm Identification algorithm when applied to the GVGAI (level 1) game-agent results dataset. }
\label{graph:a2}
\end{figure}

\begin{figure}
\centering
\includegraphics[width=1.0\linewidth]{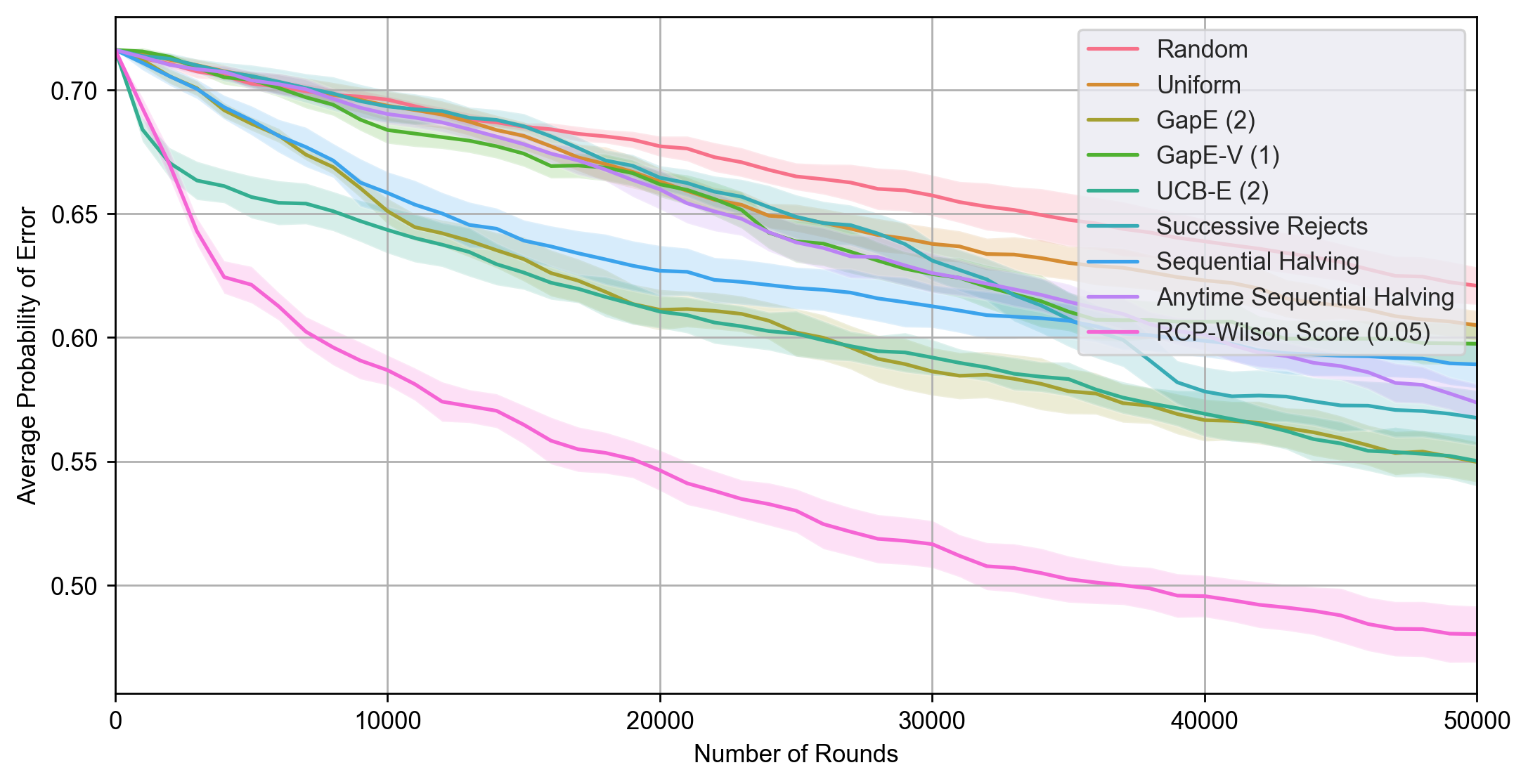}
\caption{Average probability of error for each Best Arm Identification algorithm when applied to the Ludii game-agent results dataset. }
\label{graph:a3}
\end{figure}

While the primary focus of our RCP algorithm is on minimising the average simple regret across all games, we can also measure the number of games where we correctly identify the best performing agent (average probability of error). The average probability of error of each best arm identification algorithm for the GVGAI (all levels), GVGAI (level 1) and Ludii datasets are shown in Figures \ref{graph:a1}, \ref{graph:a2} and \ref{graph:a3} respectively.

While our RCP algorithm does still outperform all other alternatives, the percentage improvement is substantially lower when measuring average probability of error (APE) compared to average simple regret. Compared to the previous second-best approach (UCB-E) RCP achieved a 10.4\% lower APE for the GVGAI (all levels) dataset, a 20.2\% lower APE for the GVGAI (level 1) dataset, and a 10.7\% lower APE for the Ludii dataset. While this performance improvement is less substantial compared to our results for average simple regret, it is still nonetheless a demonstrable reduction in APE over prior techniques.

\subsection{Uniform Bandit Selection}

To better understand the impact that our dynamic multi-bandit selection process has on the overall performance of our RCP algorithm, we also evaluated our previously described uniform bandit selection variant (Section \ref{uni_bandit}). The intention for RCP is to focus pulls on bandits where there is still a high potential for regret change, typically associated with a high uncertainty in estimating the expected reward for several top-performing arm candidates. This is in contrast to the uniform bandit selection variant, which allocates an equal number of pulls to each bandit.

\begin{figure}
\centering
\includegraphics[width=1.0\linewidth]{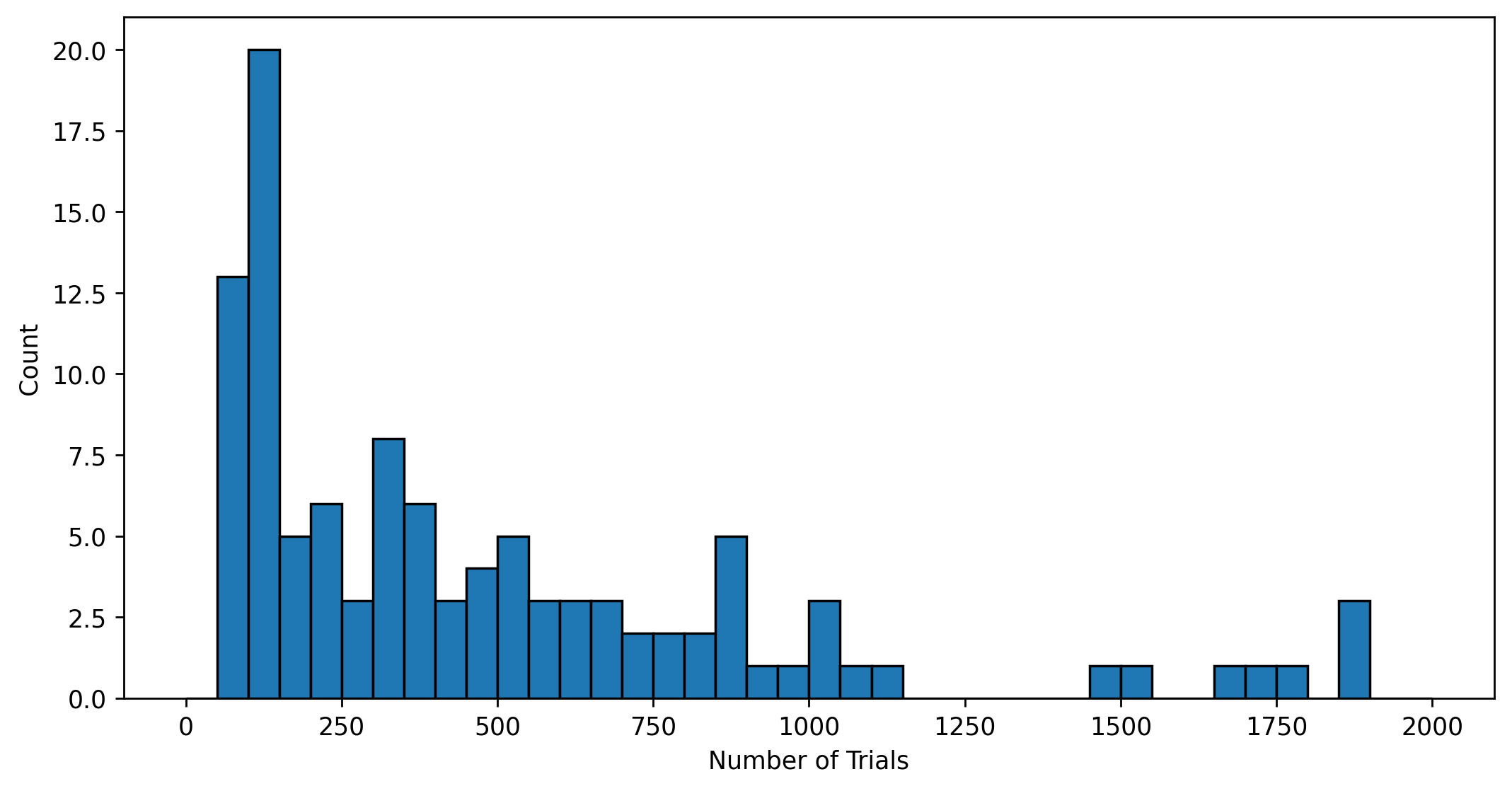}
\caption{Distribution of arm pulls (i.e., number of trials) using our RCP algorithm (Wilson score, $\alpha$=0.05) for each game in the GVGAI dataset. Uniform bandit selection allocates a fixed 462 trials per game.}
\label{graph:rcp_game_selection_gvgai}
\end{figure}

\begin{figure}
\centering
\includegraphics[width=1.0\linewidth]{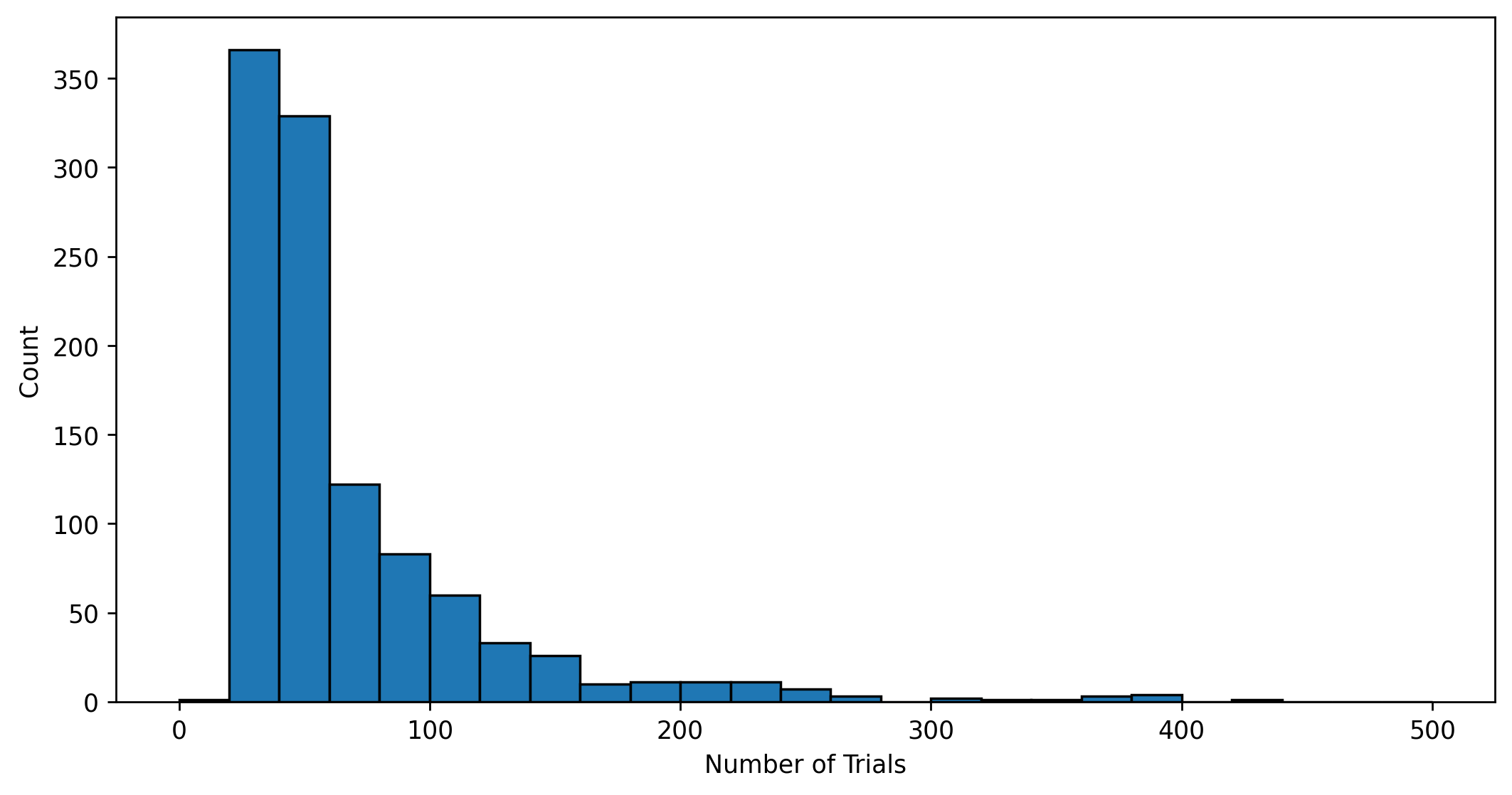}
\caption{Distribution of arm pulls (i.e., number of trials) using our RCP algorithm (Wilson score, $\alpha$=0.05) for each game in the Ludii dataset. Uniform bandit selection allocates a fixed 46 trials per game.}
\label{graph:rcp_game_selection_ludii}
\end{figure}

The regular RCP algorithm selects the arm with the highest potential of altering the regret of it's associated bandit, often leading to a highly imbalanced number of pulls for each bandit, see Figures \ref{graph:rcp_game_selection_gvgai} and \ref{graph:rcp_game_selection_ludii}. These graphs visualise the selection distribution of bandits (i.e. games) when applying the regular RCP algorithm (Wilson score, $\alpha$=0.05, averaged across all 10 runs) for each of our domains. We can see that RCP selects fewer trials for the majority of games when compared to uniform bandit selection. However, there are also several games with a comparatively large number of trails, indicating that these games had a relatively high regret change potential. 

\begin{figure}
\centering
\includegraphics[width=1.0\linewidth]{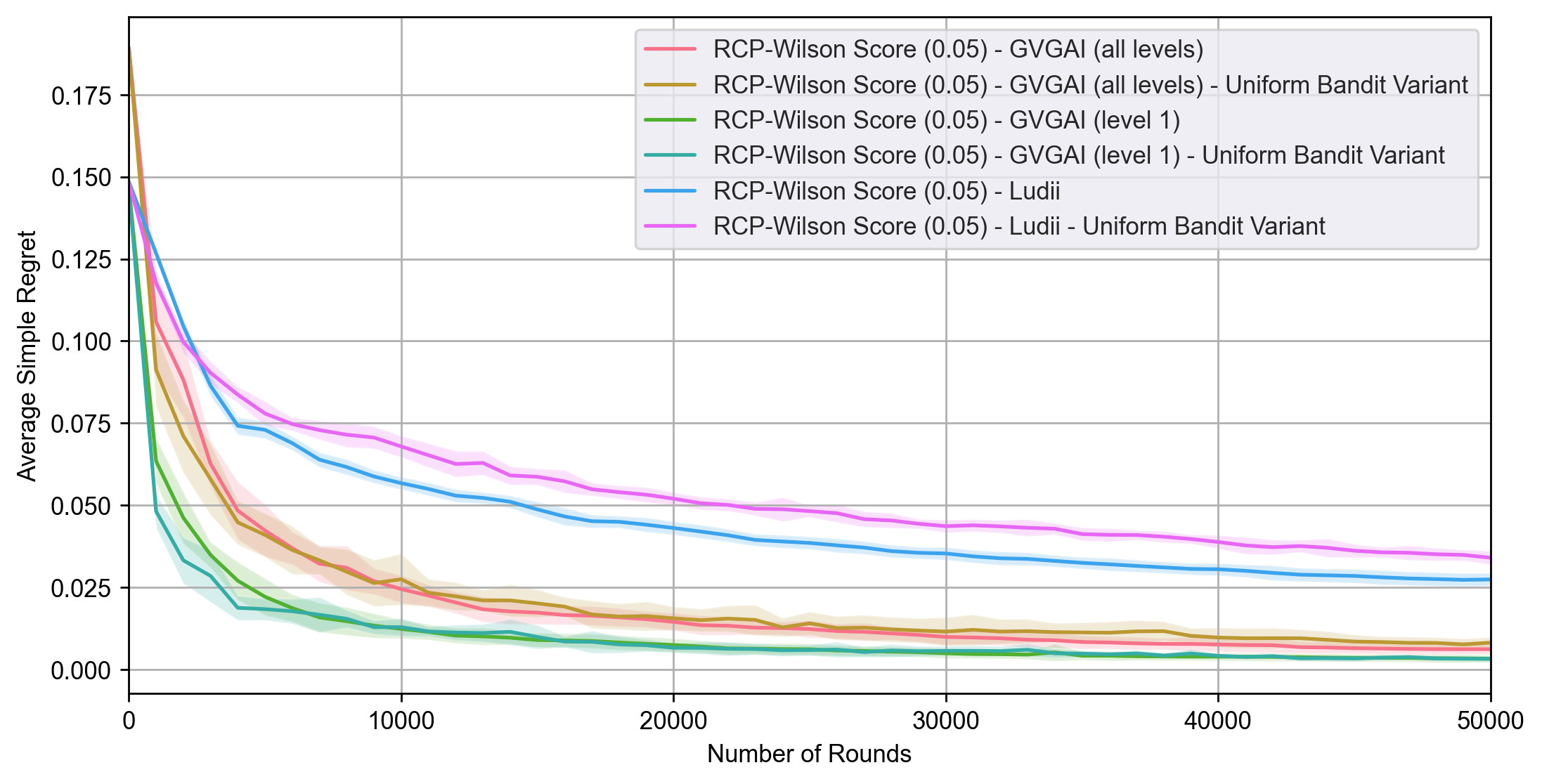}
\caption{Average simple regret for RCP (Wilson score, $\alpha$=0.05) when applied to each game-agent results dataset, using both dynamic (i.e., regular) and uniform bandit selection. The shaded uncertainty region indicates plus/minus one standard deviation.}
\label{graph:rcp_uniform_regret}
\end{figure}

Looking first at the average simple regret results, see Figure \ref{graph:rcp_uniform_regret}, we can see that the uniform bandit selection variant appears to achieve a better regret reduction when we have a small number of rounds, but that our RCP's dynamic approach to selecting which bandits to pull leads to a better long-term performance boost. This most prevalent for Ludii dataset, where it achieves an average regret 17.6\% smaller that it's uniform bandit selection variant (averaged across the full 50,000 rounds).

\begin{figure}
\centering
\includegraphics[width=1.0\linewidth]{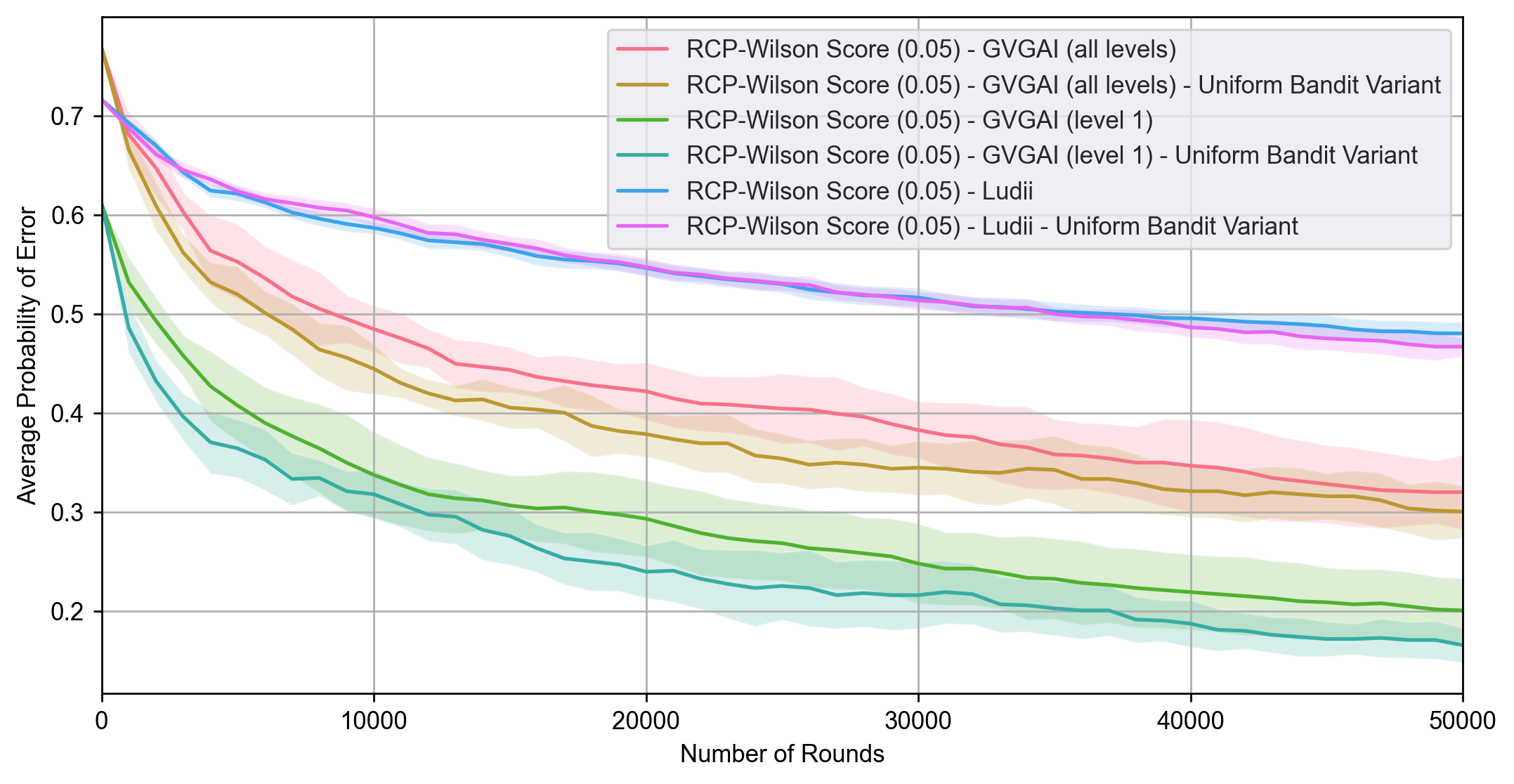}
\caption{Average probability of error for RCP (Wilson score, $\alpha$=0.05) when applied to each game-agent results dataset, using both dynamic (i.e., regular) and uniform bandit selection. The shaded uncertainty region indicates plus/minus one standard deviation.}
\label{graph:rcp_uniform_ape}
\end{figure}

In contrast, the uniform bandit selection variant seemingly performs much better when comparing the APE between the two approaches, see Figure \ref{graph:rcp_uniform_ape}. This is not unexpected, as our proposed RCP approach is designed specifically for regret reduction rather than APE, and so is likely to under-sample bandits where the selected arm is close to optimal (even if it may not actually be the best arm). However, if the goal is indeed to find the true best performing agent for each game, regardless of how small the regret difference is, then the uniform variant of RCP appears to give the best results. 

\subsection{Discussion}

Overall, the improvements in best agent identification offered by our proposed RCP algorithm are clear across both domains, being able to to consistently outperform all other techniques for the vast majority of rounds tested. We believe that one of the core reasons for this improvement is our decision to focus primarily on the simple regret metric, rather than the average probability of error (APE) metric. Many prior best-arm identification algorithms choose to focus on the later, resulting in approaches that would prioritise pulling arms in a bandit that were very near the top performing arm in a attempt to find the ``one true'' best arm. This would often result in behaviour that provides minimal benefit towards optimising simple regret. For example, in a bandit where multiple arms achieve a very high (but not perfect) success rate, it takes a lot of pulls to identify which is the best performing arm. However, from a regret metric perspective, it makes very little difference which arm we select, and our pulls would likely be more useful on other bandits.

The benefits of our RCP algorithm also appear to be most prominent in the early-mid round period, where it quickly achieved a much lower simple regret compared to other techniques. As more rounds are performed we see that this gap begins to shrink, although the overall percentage improvement remains relatively stable. Given an sufficiently large number of rounds, even random arm selection would (in theory) be able to obtain a perfect average simple regret score of zero, although such an assumption is clearly not feasible in practice. Demonstrating our RCP algorithm's improved performance over a wide range of possible rounds emphasises its practical application to estimating agent performance in a realistic experimental setting.

\section{Conclusion}
\label{sec:Conclusion}
In this paper we have presented our novel RCP algorithm for best arm identification in multi-armed bandits. This approach uses a provided confidence interval to estimate the regret change potential of each arm, and has been designed specifically for the task of identifying the best agents in a general game playing domain. Experiments conducted on two of the most popular general game systems (GVGAI and Ludii) reveal that our proposed approach significantly outperforms all previous state-of-the-art algorithms. We hope that this approach will be applied to enhance future agent evaluation procedures across a wide range of general game systems.

\subsection{Future Work}

There are several aspects that we did not consider for our presented RCP algorithm that could be explored further.
In our experiments, all trials were assumed to take an equal length of time to complete. However, this is often not the case, with certain games likely taking much longer to complete than others. Skilled agents may also be able to win easy games quicker than less skilled agents, or conversely take longer to lose at harder games. While we did not explore these aspects within this paper, it should be possible to create a ``cost aware'' version of our algorithm that considers the expected computational expense of performing a trial for a specific game-agent pair (based on prior experience).
We also did not consider any domain specific measures that could be used to identify performance correlations between games/agents. If we are able to estimate that two (or more) games/agents have similar profiles, then results obtained from one trial could potentially be applied proportionately to several other related games/agents. Domain agnostic measures of game/agent distance could also be determined based on the similarity of prior results, although these may not always prove reliable.

One limitation with some of our proposed confidence interval estimates (specifically Wilson Score, Jeffreys, Clopper–Pearson and Agresti–Coull) is that they do not consider the total number of samples (i.e., rounds) that have been performed. This makes them fast to compute, as only the confidence bounds of the last arm pulled needs to be updated, but also means that the confidence bounds of an arm only change if it is pulled. This could lead to a situation where, through very bad luck, the lower confidence bound of a non-best arm exceeds the upper confidence bound of the best arm. While this is very unlikely to happen, particularly for very small alpha values, it is still theoretically possible. One way to address this problem would be to use an adaptive alpha value that decreases as the current round number increases (similar to that of PAC). A correctly tuned adaptation rate could also lead to improved performance, by starting off with a high alpha value that finds good (although potentially not optimal) solutions quickly, and then steadily decreasing this alpha value to increase the amount of exploration (i.e., reverse annealing).

Another suitable avenue for future work would be to identify additional multi-task, multi-agent domains for carrying out further evaluations. Our approach is general enough to work on datasets for any such domains, providing the requirement on reward limits being bounded between 0 and 1 is satisfied. While we selected binomial confidence interval estimates specifically for the domain of general game playing, it is also possible that other confidence intervals would be better suited for alternative domains with different reward distributions. Further evaluations will therefore be needed to accurately estimate the wider applicability of our approach.

Lastly, one of the immediate applications of our proposed best-agent-identification process would be to improve the performance of portfolio agents that leverage a suite of other pre-existing agents to make decisions. In their simplest form, portfolio agents are classification models trained to predict the best performing agent for a specific game. These agents are trained on labelled instances of prior games where the best performing agent is already known. 
Using our proposed best-agent-identification approach, the best performing agent in any given game can be estimated far more efficiently. This can potentially lead to a significant boost in the training speed and performance of portfolio agents for general game playing, as well as other domains where ground truth labels are expensive to compute.

\appendices

\section{GVGAI (all levels) additional hyperparameter results}
\label{app:gvgai_all}
In this appendix we provide additional results for all algorithm hyperparameter values on the GVGAI (all levels) dataset.

\FloatBarrier

\section{GVGAI (level 1) additional hyperparameter results}
\label{app:gvgai_0}
In this appendix we provide additional results for all algorithm hyperparameter values on the GVGAI (level 1) dataset.

\FloatBarrier

\section{Ludii additional hyperparameter results}
\label{app:ludii}
In this appendix we provide additional results for all algorithm hyperparameter values on the Ludii dataset.

\FloatBarrier

\section{Probably Approximately Correct Result}
\label{app:pac}
We present a standard fixed-confidence guarantee for our the RCP algorithm when employing the previously described PAC confidence interval. This guarantee relies on (i) confidence intervals that hold uniformly over time and (ii) a separation-based stopping rule.

For each bandit $m\in\{1,\dots,M\}$ and arm $k\in\{1,\dots,K\}$, rewards are i.i.d.\ random variables in $[0,1]$ with mean $\mu_{m,k}$.
Recall that $T_{m,k}(t)$ denotes the number of observations collected from pair $(m,k)$ up to time $t$, and let $\hat\mu_{m,k}(t)$ be the corresponding sample mean.
For a user-defined confidence parameter $\delta\in(0,1)$, define
\begin{equation}
\delta_{m,k,t} := \frac{6\delta}{\pi^2 M K t^2},
\label{eq:delta_mkt}
\end{equation}
the confidence radius
\begin{equation}
\beta_{m,k}(t)
:=
\sqrt{
\frac{\log\!\bigl(2/\delta_{m,k,t}\bigr)}
     {2\max\{1,T_{m,k}(t)\}}
},
\label{eq:beta_mkt}
\end{equation}
and the confidence interval $[L_{m,k}(t),U_{m,k}(t)]$ with
\begin{align}
L_{m,k}(t) := \max\{0,\hat\mu_{m,k}(t)-\beta_{m,k}(t)\},\\
U_{m,k}(t) := \min\{1,\hat\mu_{m,k}(t)+\beta_{m,k}(t)\}.
\label{eq:CI_mkt}
\end{align}
Let $\hat k_m(t)\in\arg\max_k \hat\mu_{m,k}(t)$ be the estimated best arm on bandit $m$ at time $t$. We use a stopping rule where $\tau$ is the first time $t$ such that, for every $m$,
\begin{equation}
L_{m,\hat k_m(t)}(t) \ \ge\ \max_{k\neq \hat k_m(t)} U_{m,k}(t).
\label{eq:stop_rule_full}
\end{equation}
Upon stopping, we make final recommendations of $J_m := \hat k_m(\tau)$ for each $m$.

Define the event that all confidence intervals contain the true means
simultaneously for all times:
\begin{equation}
\mathcal E
:=
\bigcap_{t\ge 1}\ \bigcap_{m=1}^M\ \bigcap_{k=1}^K
\{\mu_{m,k}\in[L_{m,k}(t),U_{m,k}(t)]\}.
\label{eq:E_event}
\end{equation}

\begin{lemma}
\label{lem:uniform_coverage}
With rewards bounded within $[0,1]$, we have $\Pr(\mathcal E)\ge 1-\delta$.
\end{lemma}

\begin{IEEEproof}
Fix any triple $(m,k,t)$. The random variable $\hat\mu_{m,k}(t)$ is the average of $n$ i.i.d.\ samples in
$[0,1]$ with mean $\mu_{m,k}$. By Hoeffding's inequality, for any $n\ge 1$,
\begin{align}
\Pr\!\left(
|\hat\mu_{m,k}(t)-\mu_{m,k}| > \sqrt{\frac{\log(2/\delta_{m,k,t})}{2n}}
\ \Big|\ T_{m,k}(t)=n
\right) \notag \\
\le \delta_{m,k,t}.
\end{align}
The definition \eqref{eq:beta_mkt} uses $\max\{1,T_{m,k}(t)\}$ so the same bound
holds unconditionally, implying
\[
\Pr\big(\mu_{m,k}\notin[L_{m,k}(t),U_{m,k}(t)]\big)\le \delta_{m,k,t}.
\]
Then applying a union bound over all $m,k,t$:
\begin{align}
\Pr(\neg\mathcal E)
\le
\sum_{t=1}^\infty \sum_{m=1}^M \sum_{k=1}^K \delta_{m,k,t}
=
\sum_{t=1}^\infty MK \cdot \frac{6\delta}{\pi^2 MK t^2} \notag \\
=
\frac{6\delta}{\pi^2}\sum_{t=1}^\infty \frac{1}{t^2}
=
\frac{6\delta}{\pi^2}\cdot \frac{\pi^2}{6}
=
\delta.
\end{align}
\end{IEEEproof}

Assuming that the stopping rule has been reached after some amount of time, we have that for all bandits $m$,
\[
\mu_{m,\hat k}
\ \ge\ L_{m,\hat k}(t)
\ \ge\ \max_{k\neq \hat k} U_{m,k}(t)
\ \ge\ \max_{k\neq \hat k} \mu_{m,k}.
\]
Hence $\mu_{m,\hat k}=\max_k \mu_{m,k}$, that is, $\hat k\in\arg\max_k\mu_{m,k}$.
Therefore the probability of making at least one incorrect recommendation satisfies
\[
\Pr\big(\exists m:\ J_m \notin \arg\max_k \mu_{m,k}\big)\le \delta.
\]

\section*{Acknowledgment}
This work was supported by the European Cooperation in Science and Technology (COST) Action \#CA22145.

\printbibliography

\begin{IEEEbiography}[{\includegraphics[width=1in,height=1.25in,clip,keepaspectratio]{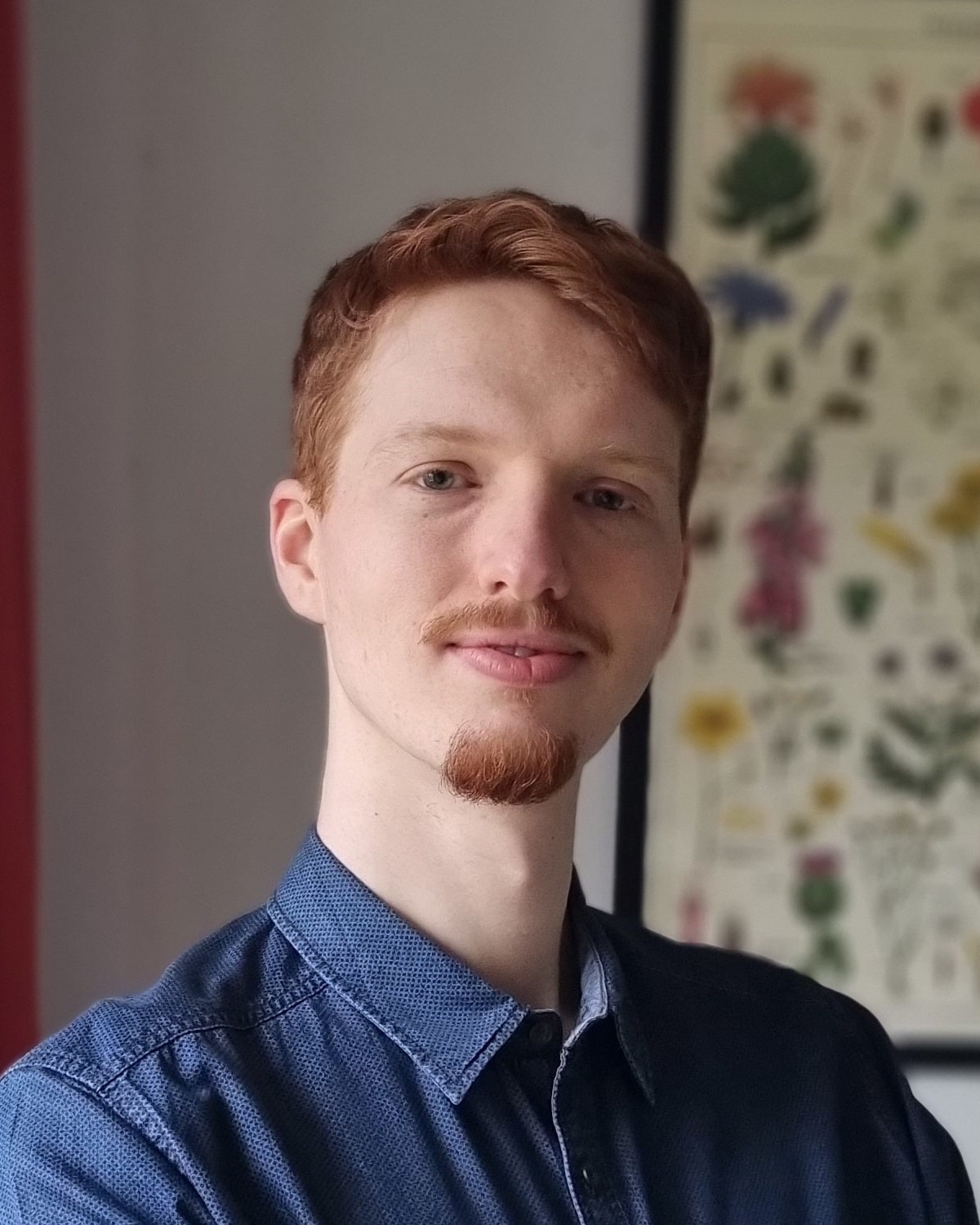}}]{Matthew Stephenson} (M'19) was born in Kent, England in 1994, before emigrating to Taranaki, New Zealand in the mid-2000s. He received a BSc (Hons) degree in Computer Science from the University of Canterbury, New Zealand in 2015, and a Ph.D. from the Australian National University (ANU) in 2019. His Ph.D. research focused on implementing various AI techniques for physics-based video games and simulations. This involved developing intelligent agents that can reason and interact within a physical environment, as well as generating content that satisfies the physical limitations of such environments.

From 2019 to 2022, he worked as a postdoctoral researcher with the Department of Advanced Computing Sciences, formally the Department of Knowledge Engineering, at Masstricht University, the Netherlands. His primary contribution during this period was on the Digital Ludeme Project, a computational study of different games throughout recorded history using modern AI techniques, hoping to chart their historical evolution and explore their role in the development of human culture. Since 2023, he has worked as a Lecturer within the College of Science and Engineering at Flinders University, South Australia. His current research focuses on applying Artificial Intelligence, Machine Learning and Data Science techniques to games. This includes designing AI to play, create and analyse games; as well as utilising games as a testbed for developing AI-based solutions to real-world problems.
\end{IEEEbiography}

\begin{IEEEbiography}[{\includegraphics[width=1in,height=1.25in,clip,keepaspectratio]{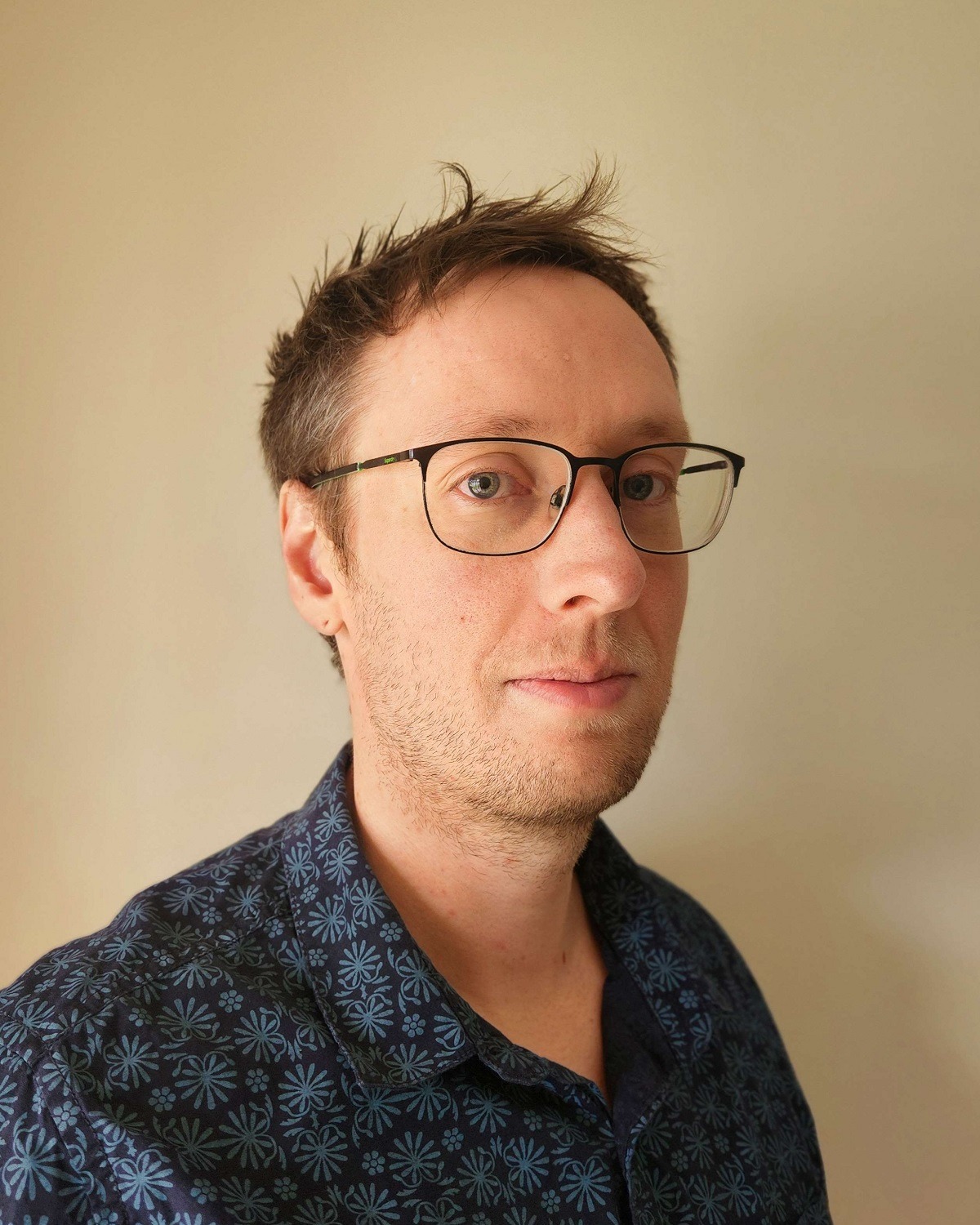}}]{Alex Newcombe} was born in Adelaide, South Australia, in 1986. He received a BSc (Hons) degree in Mathematics in 2014 and a Ph.D. degree in 2019, both from Flinders University, South Australia. His doctoral research focused on developing new computational approaches in graph theory, including the design and analysis of efficient heuristics for the visualisation of graphs.

From 2019 to 2022, he was a Postdoctoral Researcher with the Department of Mathematics, Flinders University, where he contributed to projects in both theoretical and applied mathematics, including work on graph domination, crossing numbers, and combinatorial games. Since 2022, he has been a Postdoctoral Researcher with the Centre for Defence Engineering Research and Training, engaged in multi-organisation collaborations. This research has focused on the development and implementation of geolocation algorithms for complex environments. 
\end{IEEEbiography}

\begin{IEEEbiography}[{\includegraphics[width=1in,height=1.25in,clip,keepaspectratio]{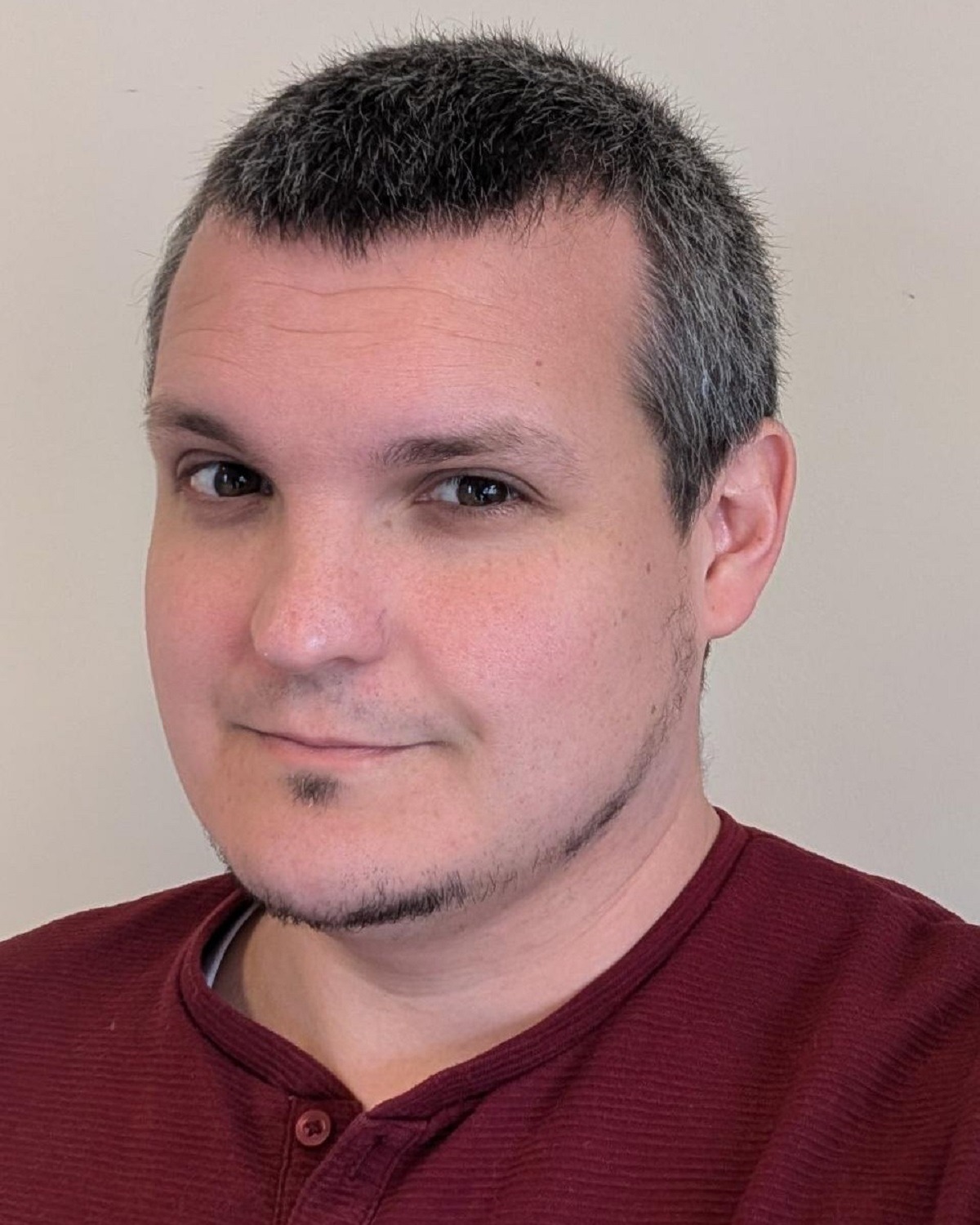}}]{{\'E}ric Piette} received his Ph.D. degree in Computer Science from Université d’Artois, France, in 2016, with a dissertation on stochastic constraint programming and bandit techniques for General Game Playing. During his Ph.D., he developed the WoodStock general game player, which won the International General Game Playing Competition in 2016 organised by the Stanford University. He was subsequently recognized with the French AI Thesis Award in 2017. From 2016 to 2018, he held Lecturer Assistant positions at Université d’Artois (CRIL) and Université Paris-Dauphine–PSL (LAMSADE). From 2018 to 2023, he was a Postdoctoral Researcher in the Department of Data Science and Knowledge Engineering (DACS), Maastricht University, contributing to the ERC-funded Digital Ludeme Project. In that context he co-developed Ludii, a ludemic general game system used to model, analyze, and play traditional board and tabletop games, and he extended his line of research in an explicitly interdisciplinary direction—using general AI approaches together with historical, archaeological, and cultural-studies methods to study games and to support the preservation of game cultural heritage through the study and reconstruction of traditional games.

Since 2023, he has been an Assistant Professor with the ICTEAM Institute and INGI Department, UCLouvain, Belgium. His research interests include Game AI, General Game Playing, human-like AI, machine learning, constraint reasoning, and knowledge representation, applied to the study and creation of strategic and traditional games. He is the main proposer and Action Chair of the COST Action GameTable (CA22145, 2023–2027), an international network on computational techniques for tabletop-game heritage. At UCLouvain, he leads the General AI Agents (GAIA) research group, advancing human-like general game-playing agents and interdisciplinary methods for the cultural heritage of games.

\end{IEEEbiography}

\begin{IEEEbiography}[{\includegraphics[width=1in,height=1.25in,clip,keepaspectratio]{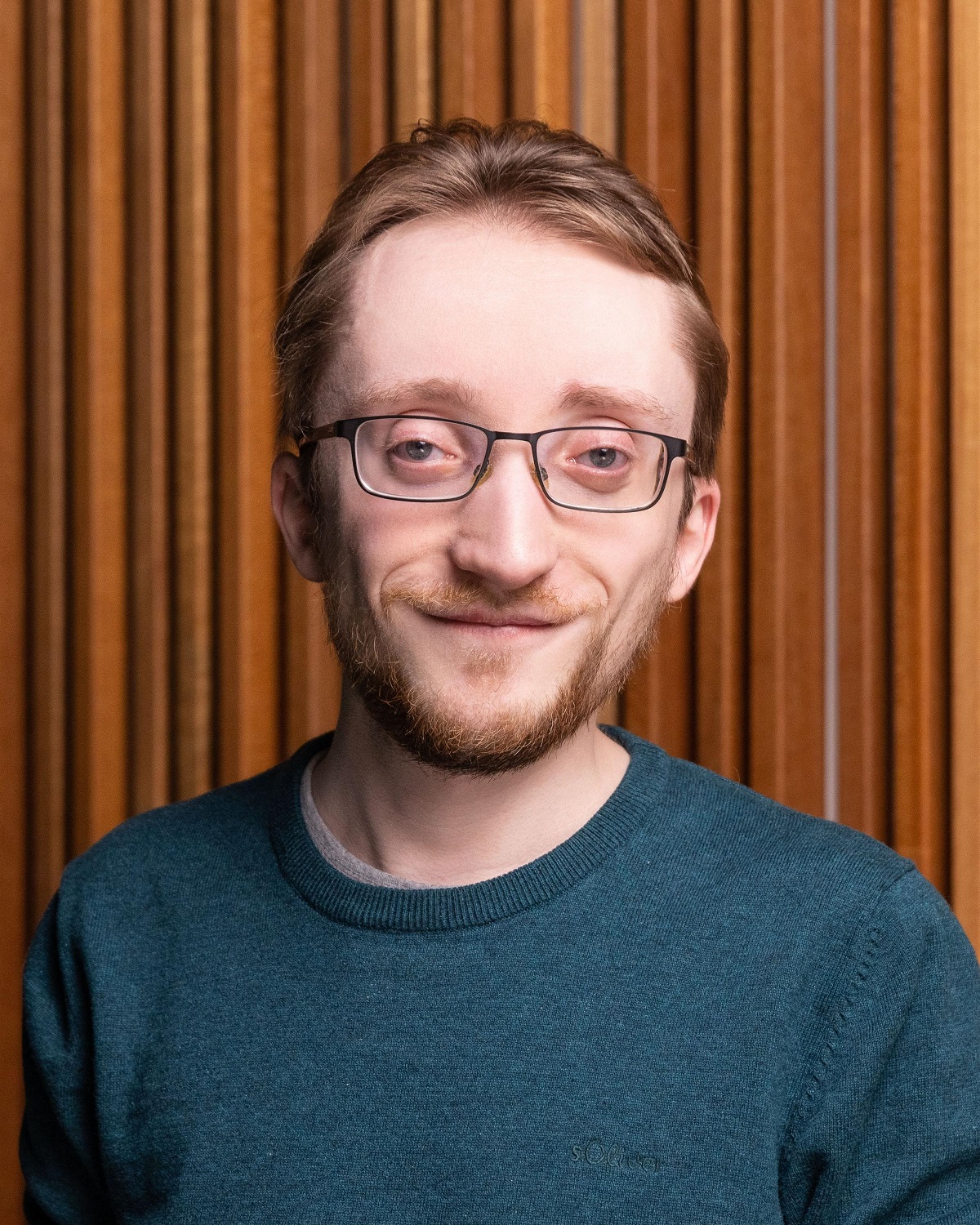}}]{Dennis Soemers} (M'19) was born in Maastricht, the Netherlands, in 1993. He received his M.Sc. degree in Artificial Intelligence from Maastricht University in 2016. He worked at Vrije Universiteit Brussel until 2018, and then continued at Maastricht University. In the fall of 2020, Dennis worked as a Research Intern at Facebook AI Research. He received his Ph.D. degree from Maastricht University in 2023, where he continued working as a Postdoc and transitioned into the role of Assistant Professor in September 2024. His primary research interests are reinforcement learning, search algorithms, artificial intelligence (AI) for game playing, and other applications of AI to sequential decision-making processes and adversarial settings.
\end{IEEEbiography}

\EOD

\end{document}